%% file: HINT.tex
\documentclass[10pt,twocolumn,letterpaper]{article}

\usepackage[pagenumbers]{iccv} 


\definecolor{iccvblue}{rgb}{0.21,0.49,0.74}
\usepackage[pagebackref,breaklinks,colorlinks,allcolors=iccvblue]{hyperref}
\usepackage{adjustbox}
\usepackage{multirow}

\def\paperID{12111} 
\def\confName{ICCV}
\def\confYear{2025}

\title{Devil is in the Uniformity: Exploring Diverse Learners within Transformer for Image Restoration}

\author{
~~Shihao Zhou$^{1,2}$~~~~~~~Dayu Li$^1$~~~~~~~~Jinshan Pan$^3$~~~~~~Juncheng Zhou$^1$~~~~~~Jinglei Shi$^1$~~~~~~
Jufeng Yang$^{1,2}$\\
$^1$ VCIP \& TMCC \& DISSec, College of Computer Science, Nankai University\\
$^2$ Nankai International Advanced Research Institute (SHENZHEN· FUTIAN)\\
$^3$ School of Computer Science and Engineering, Nanjing University of Science and Technology\\
\hspace{-8pt}
{\tt\small{zhoushihao96@mail.nankai.edu.cn, ldy030911@gmail.com, sdluran@gmail.com}}\\
\hspace{-6pt}
{\tt\small{ 2112612@mail.nankai.edu.cn, jinglei.shi@nankai.edu.cn, yangjufeng@nankai.edu.cn}}
\vspace{-5mm}
}

\begin{document}
\maketitle

\begin{abstract}
%
Transformer-based approaches have gained significant attention in image restoration, where the core component, i.e, Multi-Head Attention (MHA), plays a crucial role in capturing diverse features and recovering high-quality results. 
In MHA, heads perform attention calculation independently from uniform split subspaces, and a redundancy issue is triggered to hinder the model from achieving satisfactory outputs. 
%
In this paper, we propose to improve MHA by exploring diverse learners and introducing various interactions between heads, which results in a \textbf{H}ierarchical mult\textbf{I}-head atte\textbf{N}tion driven \textbf{T}ransformer model, termed HINT, for image restoration. 
%
%
HINT contains two modules, i.e., the Hierarchical Multi-Head Attention (HMHA) and the Query-Key Cache Updating (QKCU) module, to address the redundancy problem that is rooted in vanilla MHA. 
Specifically, HMHA extracts diverse contextual features by employing heads to learn from subspaces of varying sizes and containing different information. 
Moreover, QKCU, comprising intra- and inter-layer schemes, further reduces the redundancy problem by facilitating enhanced interactions between attention heads within and across layers. 
Extensive experiments are conducted on \textbf{12} benchmarks across \textbf{5} image restoration tasks, including low-light enhancement, dehazing, desnowing, denoising, and deraining, to demonstrate the superiority of HINT. 
\textbf{The source code will be available at \href{https://github.com/joshyZhou/HINT}{https://github.com/joshyZhou/HINT}}.

\end{abstract}

\input{sec/1_intro}
\input{sec/2_relate}

\input{sec/3_method}

\input{sec/4_exp}

\input{sec/5_con}
{
    \small
    \bibliographystyle{ieeenat_fullname}
    \bibliography{HINT}
}

\end{document}

%% file: sec/1_intro.tex
\section{Introduction}
\label{sec:intro}
\begin{figure}[t!]
\centering
\includegraphics[width=.98\linewidth]{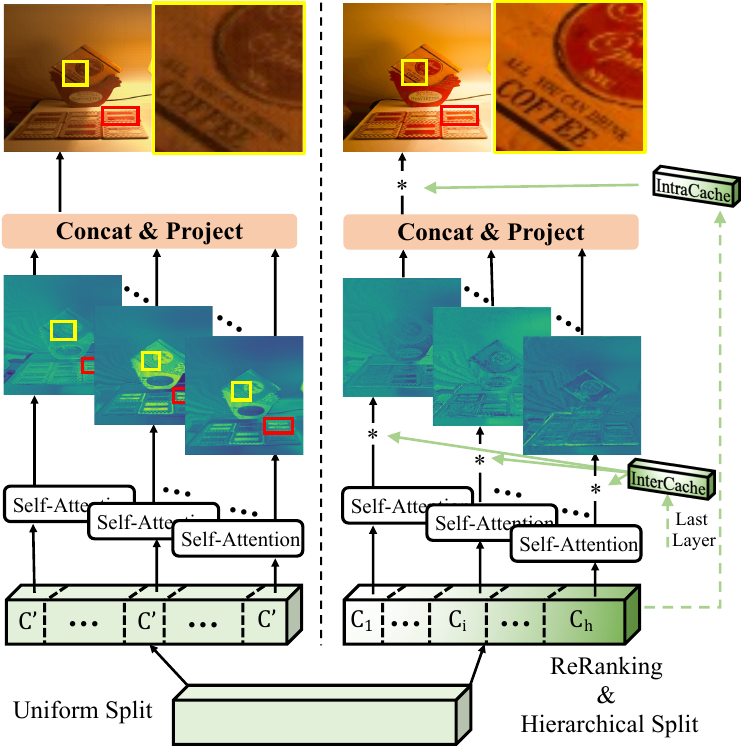}
\vspace{-3mm}
\caption{ 
Comparisons between the vanilla MHA~\cite{zamir2022restormer,iccv2021_swinIR,retinexformer} (\textit{Left}) and the proposed HMHA equipped with the QKCU module (\textit{Right}), for the low-light enhancement task. 
The standard MHA assigns $h$ heads with subspaces of the same size ($C'$), and each head performs attention calculation independently. 
As a result, these heads intend to focus on the same regions (\textcolor{red}{red} boxes) and neglect the restoration of some degraded areas (\textcolor{yellow}{yellow} boxes), leading to an unsatisfactory output (losing details and introducing blur effect). 
In contrast, HMHA implements the reranking operation before a hierarchical subspace split, which encourages the model to learn diverse representative features. 
The QKCU enhances interactions between heads via intra-/inter-layer ways, modulating predicted features in HMHA and leading to better outputs. 
}
\label{pic:motivation}
\vspace{-7mm}
\end{figure} 
The Transformer-based frameworks~\cite{vaswani2017attention,iclr2021_vit} achieve promising performance in the field of image restoration. 
The success of this paradigm is rooted in the self-attention mechanism (SA) modeling non-local relations of pixels, which is crucial for recovering the global structure of the image.
Many works pay attention to develop efficient SA variants for high-quality outputs~\cite{zamir2022restormer,iccv2021_swinIR,eccv2022_Stripformer}. 
While it is worth noting that multi-head attention (MHA), employing multiple heads to perform self-attention computation in parallel from uniform split subspace, serves as the key fundamental component in embracing high computational efficiency and enhancing the diversity of the captured features. 

The conventional MHA mechanism has a drawback of the redundancy issue. 
Some researches~\cite{wang2022improved,nguyen2022improving,MGK_icml22}, in the field of NLP, have pointed out that specialized heads contribute most to the final decision while others can be pruned. 
In this paper, we demonstrate that this problem exists in the restoration tasks, and further find out that the redundancy issue can be traced to the focus on the same region by different heads. 
As shown in Figure~\ref{pic:motivation}, the standard MHA mechanism assigns $h$ heads with subspaces of the same dimension ($C'$), where each head performs attention calculations in parallel and independently of the others. 
The visualized features from different heads disclose the problem of the vanilla MHA - that is, the inclination to pay attention to the same region (red boxes) which is redundant, and neglect of restoration on some degraded areas (yellow boxes) which causes unpleasant outputs. 

To address this limitation, we improve the MHA from two perspectives. 
\textbf{First}, 
the heads learn from subspaces of uniform size, which contain similar information, leading to the problem of focusing on the same regions. 
%
Intuitively, to mitigate this issue, we introduce a ranking paradigm in terms of channel similarity before a hierarchical subspace splitting. 
In this way, each subspace contains information independent from others, and their sizes are different. 
This design explores heads as diverse learners, resulting in a hierarchical Multi-Head Attention module, namely HMHA, to replace the vanilla counterpart. 
It is capable of extracting diverse representative features, compared to the conventional MHA. 
\textbf{Second}, a lack of collaboration between heads makes the redundancy problem worse. 
We propose to enhance interactions between heads in intra- and inter-layer schemes through a Query-Key Cache Updating (QKCU) mechanism. 
Specifically, the intra-layer cache serves as a gating module, enhancing useful information in aggregated features that are captured by heads. 
%
%
On the other hand, the inter-layer cache emphasizes modulating the calculated attention score of each head, using history attention scores. 
Both intra- and inter- modulation are dependent on inputs, which improve the capability of HMHA to learn diverse contextual representations.
Building upon the two key components, i.e., HMHA and QKCU, we propose {HINT}, a \textbf{H}ierarchical multi-head atte\textbf{N}tion driven \textbf{T}ransformer model for image restoration. 
%

Overall, we summarize the contributions of this work:
\begin{itemize}
  \item  
  We present HINT, a \textbf{H}ierarchical multi-head atte\textbf{N}tion driven \textbf{T}ransformer model for removing undesired degradations from images. 
  HINT demonstrates the effect of exploring diverse learners in Multi-Head Attention (MHA) and enhancing interactions via inter- and intra-layer ways for the problem of image restoration.
  \item HINT introduces Hierarchical Multi-Head Attention (HMHA), which alleviates the redundancy issue in standard MHA by enabling the model to learn distinctive contextual features from different subspaces. 
  Additionally, HINT incorporates the Query-Key Cache Updating (QKCU) mechanism, combining intra- and inter-layer schemes to enhance interaction between heads. 
  \item Extensive qualitative and quantitative evaluations are conducted on \textbf{12} benchmarks for \textbf{5} typical image restoration tasks: low-light enhancement, dehazing, desnowing, denoising, and deraining, where HINT performs favorably against state-of-the-art algorithms in terms of restored image quality and model complexity.
  \end{itemize}


%% file: sec/2_relate.tex
\section{Related Work}
\label{sec:relatedWork}

\subsection{Image Restoration} 
Capturing images in unsatisfactory environments often results in low-quality outputs, negatively impacting downstream tasks~\cite{ChenJPSWY24,DuanBXQHT24,0001HJH24}. 
Image restoration offers a plausible solution by recovering clear images from undesired degradations, e.g., haze~\cite{he2010single,li2017aod,10076399}, rain streaks~\cite{10035447,purohit2021spatially,wang2019spatial}, and low-light~\cite{retinexformer,Chen2018Retinex,xu2022snr}. 
Over the past decades, the research community has witnessed the paradigm shift from conventional hand-crafted approaches~\cite{cho2009fast,fergus2006removing} to learning-based CNN models~\cite{yuke_pami22_car_blur_noise,liu2018non,nips21_luo_sr}. 
To achieve improved restoration performance, various efficient modules and advanced architecture designs have been proposed. 
Among these, residual feature learning~\cite{liu2019dual,zhang2018residual,gu2019self} with skip connection, encoder-decoder architecture~\cite{cvpr2021cho,kupyn2019deblurgan,yue2020dual} for hierarchical representations, and attention mechanisms~\cite{MIRNetv2,pami21_deng,tip23_songxibin} to emphasize important signals have become the popular ingredients. 
%

In recent years, Transformer-based models~\cite{vaswani2017attention} have been adapted for low-level tasks, achieving significant advancements across various image restoration tasks~\cite{retinexformer,xiao2022image,eccv2022_Stripformer}. 
IPT~\cite{chen2021pre} is a pioneering work that applies the vision Transformer~\cite{iclr2021_vit} to low-level tasks, and obtains surprising results. 
The quadratic complexity of vanilla self-attention hinders it from applying high-resolution inputs, prompting researchers to explore solutions for reducing computational loads. 
To address this, Restormer~\cite{zamir2022restormer} computes attention scores along channel dimension. 
Another approach is window-based attention~\cite{liu2021swin}, which is adopted by models such as Uformer~\cite{wang2022uformer} and SwinIR~\cite{iccv2021_swinIR}. 
Although these attention mechanisms successfully alleviate the computational burden, they still rely on the vanilla MHA~\cite{vaswani2017attention}, which incurs redundancy issue~\cite{MGK_icml22,16better1_nips19,xiao2024improving} and further limits the representation capacity of the model.

\begin{figure*}[htp!]
\centering
\includegraphics[width=\linewidth]{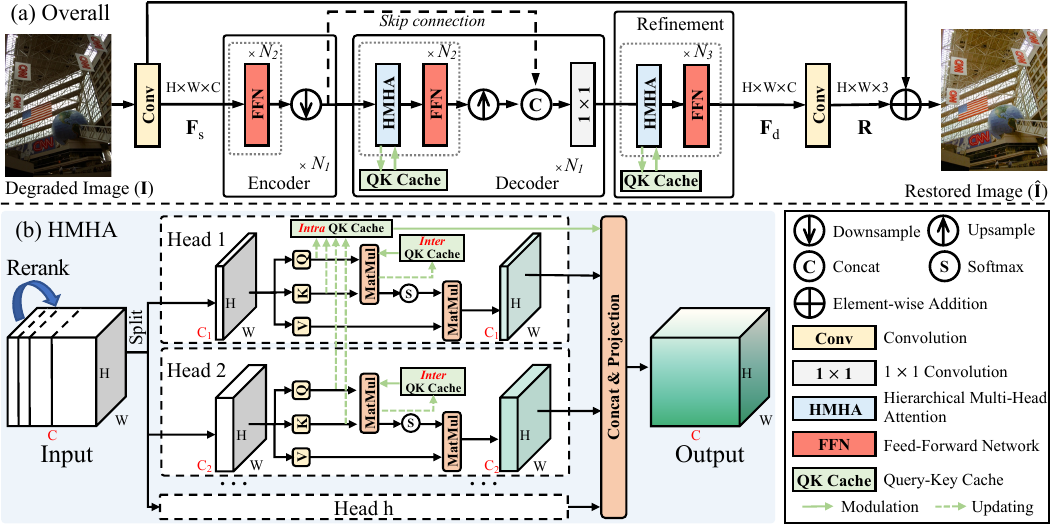}
\vspace{-7mm}
\caption{Illustration of the proposed \textbf{H}ierarchical multi-head atte\textbf{N}tion driven \textbf{T}ransformer model ({HINT}). 
(a) Overview architecture of the proposed HINT. 
(b) Hierarchical Multi-Head Attention (HMHA) mechanism.
}
\label{pic:overall}
\vspace{-6mm}
\end{figure*}

\subsection{Multi-Head Attention}
As a fundamental component of the Transformer, MHA plays a crucial role in capturing diverse relationships and achieving impressive performance in practice. 
Unfortunately, it is well known in the research community that not all heads contribute equally to the given task~\cite{acl_voita2019analyzing}. 
Previous works have attempted to address this by incorporating interaction or collaboration between heads~\cite{ahmed2017weighted,acl_li2019information,emnlp_2022_mixture}, however, the expressive power gained from these methods remains limited, as the heads still operate independently~\cite{xiao2024improving}. 
Another potential remedy is to modulate the query, key, and value projections of heads~\cite{cordonnier2020multi,liu2022tuformer}. 
While this modifies the underlying information flow, it lacks adaptability to inputs due to its static nature~\cite{xiao2024improving}. 
More recently, the community has explored dynamically modulating attention scores~\cite{shazeer2020talking,wang2022improved,nguyen2022improving}, which ensures improvement of expressive power without introducing much computational loads. 

In this study, towards solving the image restoration problem, we build upon the basic idea of modulating attention scores~\cite{nguyen2022improving,xiao2024improving}, while further mitigating the limited express power issue that is rooted in the vanilla MHA. 
Specifically, standard MHA assigns each head with subspaces using the same dimension size~\cite{zamir2022restormer,chen2021pre,iccv2021_swinIR}, limiting the ability of the heads to learn distinctive features and resulting in redundancy. 
In contrast, we propose to learn hierarchical representations using HMHA that extracts diverse contextual information from heads using different dimensional subspaces. 
Additionally, we introduce the QKCU mechanism, incorporating both intra- and inter-layer schemes, to enhance interaction among heads. 
Unlike existing restoration methods to implement modulation of attention scores in a static projection way~\cite{zhou2024AST,DRSformer}, HINT makes the modulation dynamic, achieving better model expressiveness.

%% file: sec/3_method.tex
\section{Method}
\label{sec:method}
\noindent\textbf{Overall Pipeline.}
As illustrated in Figure~\ref{pic:overall}, the proposed HINT follows an encoder-decoder architecture. 
Given a degraded input $\mathbf{I}\in\mathbb{R}^{{H}\times{W}\times3}$, a convolutional layer is first used to extract the shallow feature $\mathbf{F}_s\in\mathbb{R}^{{H}\times{W}\times{C}}$, where $H$, ${W}$, and $C$ represent the height, width and channel dimension, respectively. 
The shallow feature is then processed through the $N_1$-level restoration pipeline to produce a deep feature $\mathbf{F}_d\in\mathbb{R}^{{H}\times{W}\times{C}}$. 
At each level of both the encoder and decoder, there are $N_2$ basic blocks, along with a convolution layer for down-sampling or up-sampling. 
Following prior works~\cite{zhou2024AST,kong2023efficient,zhou2025seeing}, we adopt an asymmetric design, which omits the self-attention mechanism in the encoder part, to boost the restoration performance. 
Thus, the basic block in the encoder consists only of a Feed-Forward Network (FFN)~\cite{zamir2022restormer}, while in the decoder, the block includes both the proposed HMHA and an FFN. 
We employ skip connection operation with $1\times{1}$ convolution to take advantage of features from the encoder in the decoder layers. 
After that, a refinement stage, consisting of $N_3$ basic blocks, is developed to further enhance the learned features.
Finally, a $3\times{3}$ convolution layer processes the deep feature $\mathbf{F}_d$ to generate the residual image $\mathbf{R}\in\mathbb{R}^{{H}\times{W}\times{3}}$. 
The restored image is obtained by adding the residual image to the degraded one, i.e., $\mathbf{\hat{I}} = \mathbf{I} + \mathbf{R}$.

\subsection{Hierarchical Multi-head Attention}
\label{subsec:HMSA}
Standard MHA assigns each head with a subspace of the same size containing similar information, which hampers the ability to learn distinctive features, leading to redundancy. 
To address this, HINT forms the HMHA to explore different dimensional subspaces, encouraging each head to learn diverse contextual information. 

We begin by revisiting the scaled dot-product attention mechanism in MHA~\cite{vaswani2017attention}, which is adopted in mainstream approaches~\cite{zamir2022restormer,iccv2021_swinIR,zhou2024AST}. 
Given a normalized input tensor $\mathbf{X} \in \mathbb{R}^{H\times W\times C}$, the attention score is calculated as:
\begin{equation}
\begin{aligned}
\text{Attention(Q,K,V)} &= \text{Softmax}\left(\frac{\mathbf{Q}\mathbf{K}^T}{\sqrt{d_k}}\right)\mathbf{V},\\
\mathbf{Q} = \mathbf{XW}_Q, \mathbf{K} &= \mathbf{XW}_K, \mathbf{V} = \mathbf{XW}_V,
\label{eq:dot_product}
\end{aligned}
\end{equation}
where $\mathbf{W}_Q\in\mathbb{R}^{C\times d_k},\mathbf{W}_K\in\mathbb{R}^{C\times d_k}, \text{and~} \mathbf{W}_V\in\mathbb{R}^{C\times d_v}$ are the linear projection matrices for the query ($\mathbf{Q}$), key ($\mathbf{K}$), and value ($\mathbf{V}$). 
Notably, for self-attention, the dimensions of key and value are equal, i.e., $d_k = d_v$. 

The conventional MHA~\cite{iclr2021_vit,zamir2022restormer} leverages multiple heads to perform scaled dot-product attention in parallel. 
By operating the attention function on different subspaces, the representation power of the model is supposed to be enhanced. 
Specifically, standard MHA adopts $h$ heads to learn representations of ($\mathbf{Q,K,V}$) in different subspace, i.e, $d_k/h$. 
The MHA can be formally expressed as:
\begin{equation}
\begin{aligned}
\text{MultiHead(X)} &= \text{Concat}(\mathrm{H_{1}},\mathrm{H_{2}},...,\mathrm{H_{h}})\mathbf{W}_p,\\
\mathrm{H_{i}} &= \text{Attention}(\mathbf{X}\mathbf{W}_{Q}^{i},\mathbf{X}\mathbf{W}_{K}^{i},\mathbf{X}\mathbf{W}_{V}^{i}),
\label{eq:dot_product_mha}
\end{aligned}
\end{equation}
where $\mathbf{W}_{Q}^{i}\in\mathbb{R}^{C\times d_k/h},\mathbf{W}_{K}^{i}\in\mathbb{R}^{C\times d_k/h}, \text{and~} \mathbf{W}_{V}^{i}\in\mathbb{R}^{C\times d_v/h}$ are the projection matrices for the $i$-th head. The output projection matrix $\mathbf{W}_p \in \mathbb{R}^{d_v\times d_{out}}$ aggregates the features captured by all heads. 

To enhance the express power of MHA, we introduce a hierarchical representation learning process. 
To be specific, the proposed HMHA assign each head to a different dimensional subspace by splitting the channel space as $C = [C_1,C_2,\dots,C_h]$ with $C_1\leq C_2\leq ...\leq C_h$. 
Before performing this split, we first rerank the channels based on their similarity to ensure that each head focuses on distinct semantic features.
Each head then operates within its own subspace, performing dot-product attention in different subspace to capture diverse contextual information.

\begin{figure}[tp!]
\footnotesize
\centering
\begin{tabular}{ccc}
\begin{adjustbox}{valign=t}
\begin{tabular}{cc}
\includegraphics[width=0.18\textwidth,height=7cm]{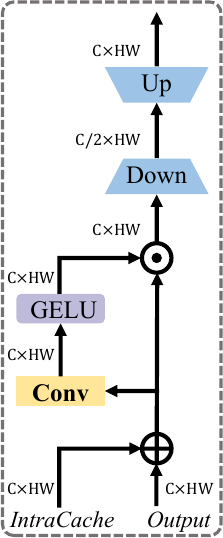}  \hspace{-1mm} &
\includegraphics[width=0.18\textwidth,height=7cm]{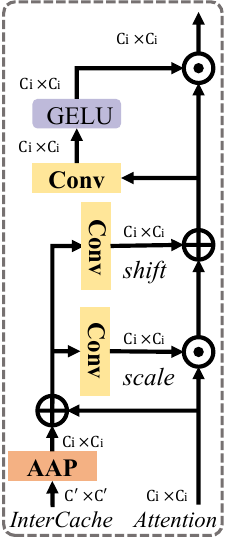} \hspace{-1mm} 
\\
(a) Intra Modulation \hspace{-4mm} &
(b) Inter Modulation \hspace{-4mm} 
\\
\end{tabular}
\end{adjustbox}
\end{tabular}
\vspace{-1mm}
\caption{Query-Key Cache Updating Mecanism. 
%
} 
\label{pic:CacheUpdating}
\vspace{-3mm}
\end{figure}

\subsection{Query-Key Cache Updating Mechanism}
\label{subsec:QKcache}
The HMHA mechanism encourages multiple heads to learn features containing diverse contextual information, while these heads still work independently as conventional works~\cite{iccv2021_swinIR,zamir2022restormer}. 
In this way, there is a lack of collaboration within the layer and across the entire model. 
As a result, the model is hindered from achieving optimal restoration performance. 
%
To mitigate this issue, we develop a QKCU mechanism that enhances interaction among heads, both within layers and across layers, as illustrated in Figure~\ref{pic:CacheUpdating}.

\noindent\textbf{Intra Query-Key Cache Modulation \& Updating.}
Given two input tensors, including the IntraCache $\mathbf{F_{intra}}\in\mathbb{R}^{C\times HW}$ and Output of HMHA $\mathbf{F_{out}}\in\mathbb{R}^{C\times HW}$, we begin by computing the sum of the two features $\mathbf{F^{s}_{intra}}=\mathbf{F_{intra}}+\mathbf{F_{out}}$. 
To selectively retain the most informative elements within the flow of information, we introduce a gating mechanism, which can be formulated as:
\begin{equation}
\begin{aligned}
\mathbf{F_{gated}} = \text{GELU}(\text{Conv}(\mathbf{F^{s}_{intra}}))\odot\mathbf{F^{s}_{intra}},
\label{eq:gatingIntraCache}
\end{aligned}
\end{equation}
where $\mathbf{F_{gated}}\in\mathbb{R}^{C\times HW}$ is the result of the gating mechanism, $\odot$ denotes element-wise multiplication, ${\rm Conv}(\cdot)$ and ${\rm GELU}(\cdot)$ are convolution operation and the GELU activation function~\cite{hendrycks2016gaussian}, respectively. 
Then, we adapt this transformed feature via feature compression and rebuilding:
\begin{equation}
\begin{aligned}
\mathbf{F^{o}_{Intra}} =\rm{Conv_{up}(}\rm {Conv_{down}(} \mathbf{F_{gated}}\rm{))},
\label{eq:LRA_IntraCache}
\end{aligned}
\end{equation}
where $\mathbf{F^{o}_{Intra}}\in\mathbb{R}^{C\times HW}$ is the output feature after the modulation process, $\rm{Conv_{up}(\cdot)}$ and  $\rm{Conv_{down}(\cdot)}$ are convolution operations that project the channel dimension into higher and lower levels, respectively. 
This design encourages the model to focus on the key features of the data. 

Next, to update the Intra Query-Key Cache in each layer, we compute the sum of the Query ($\mathbf{Q}$) and Key ($\mathbf{K}$) projections for each head as follows:
\begin{equation}    
\begin{aligned}
\mathbf{F}^{i} &=\mathbf{Q}^{i}+\mathbf{K}^{i},\\
\mathbf{F_{intra}} &= \rm{Concat(\mathbf{F}^{1},...,\mathbf{F}^{h})}
\label{eq:update_IntraCache}
\end{aligned}
\end{equation}

\noindent\textbf{Inter Query-Key Cache Modulation \& Updating.} 
Given two input tensors, the InterCache $\mathbf{F_{inter}}\in\mathbb{R}^{C'\times C'}$ and the dot-product attention of query and key $\mathbf{F_{att}}\in\mathbb{R}^{C_i\times C_i}$, we first resize $\mathbf{F_{inter}}$ to obtain $\mathbf{\hat{F}_{inter}}\in\mathbb{R}^{C_i\times C_i}$. 
After that, we compute the modulation components by summing $\mathbf{F_{att}}$ and the resized feature map $\mathbf{\hat{F}_{inter}}$, resulting in: $\mathbf{F^{s}_{inter}}=\mathbf{F_{att}}+\mathbf{\hat{F}_{inter}}$. 
We then calculate `scale' and `shift' components for pixel-wise modulation:
\begin{equation}
\begin{aligned}
\mathbf{F_{shift}} &= \mathbf{F^{s}_{inter}}\mathbf{W_{shift}},
\mathbf{F_{scale}} = \mathbf{F^{s}_{inter}}\mathbf{W_{scale}},\\
\mathbf{F_{m}}&=\mathbf{F_{scale}}\odot\mathbf{F_{att}}+\mathbf{F_{shift}},
\label{eq:component_InterCache}
\end{aligned}
\end{equation}
where $\mathbf{F_{m}}\in\mathbb{R}^{C_i\times C_i}$ is the modified feature map , $\mathbf{W_{scale}}$ and $\mathbf{W_{shift}}$ are projection matrices that learned for $\mathbf{F_{scale}}\in \mathbb{R}^{C_i\times C_i}$ and $\mathbf{F_{shift}}\in \mathbb{R}^{C_i\times C_i}$ components, respectively. 
Similarly, this feature is further transformed via a gating mechanism, which can be defined as:
\begin{equation}
\begin{aligned}
\mathbf{F^{o}_{inter}} = \text{GELU}(\text{Conv}(\mathbf{F_{m}}))\odot\mathbf{F_{m}},
\label{eq:gatingInteraCache}
\end{aligned}
\end{equation}
where $\mathbf{F^{o}_{inter}}\in \mathbb{R}^{C_i\times C_i}$ is the output feature. 

Next, to update the Inter Query-Key Cache across layers, we implement a layer-wise cache calculation process and refine the historical cache. The cache for the current layer: 
\begin{equation}
\begin{aligned}
\mathbf{F}^{l}_{\mathbf{inter}} =  \sum_{i=1}^{h} {\rm R}(\mathbf{
Q}^{i}{\mathbf{K}^{i}}^\mathit{T}){C_i/C}	,
\label{eq:layerCalInteraCache}
\end{aligned}
\end{equation}
where R($\cdot$) is a function that resizes the feature map of each head to a uniform shape, and $C_i/C$ is the assigned weight of each feature, with $C$ being the total number of channels. 
Building upon the cache of the current layer, we progressively update the inter-layer cache as follows:
\begin{equation}
\begin{aligned}
\mathbf{F_{inter}} = \alpha\mathbf{F_{inter}} + (1-\alpha)\mathbf{F}^{l}_{\mathbf{inter}},
\label{eq:crosslayerCalInteraCache}
\end{aligned}
\end{equation}
where the hyper-parameter $\alpha$ controls the degree of information flow within the inter-layer cache.

%% file: sec/4_exp.tex
\section{Experiments}
\label{sec:exp}
In this section, we evaluate HINT on \textbf{12} benchmark datasets across \textbf{5} typical image restoration tasks, including low-light enhancement, dehazing, and desnowing, denoising, and deraining. 
Due to the limited space, additional results (image deraining and denoising) and detailed experimental settings are provided in the supplementary material. 

\subsection{Experimental Settings}
\label{subsec:ExpSet}
\noindent\textbf{Implementation Details.}
HINT consists of an encoder-decoder with $N_1$ = 4 levels, where both the encoder and decoder share the same block structure: $N_2$ = [4, 6, 6, 6]. 
At the 4-th level, the encoder and decoder block are unified into a bottleneck layer, following the design of~\cite{wang2022uformer}. 
The refinement stage contains $N_3$= 4 blocks, and the embedding dimension $C$ is set as 48. 
The number of attention heads is 4, with the dimensional ratio of [1, 2, 2, 3]. 
%
%
The reranking strategy adopted in HINT is built on Pearson correlation-based similarity.
The hyper-parameter $\alpha$ in Equation~\eqref{eq:crosslayerCalInteraCache} is experimentally set to 0.9.
The AdamW optimizer is adopted to train HINT. 
The widely adopted loss functions~\cite{wang2023promptrestorer,zhou2025seeing} are employed to constrain the model training.

\begin{table}[!tp]
\caption{Quantitative comparisons on LOL-v2~\cite{yang2021sparse} for low-light enhancement. 
* indicates unsupervised methods.
}
\vspace{-3mm}
\centering
\footnotesize
\setlength{\tabcolsep}{.5mm}
{
\begin{tabular}{cccccccccc}
\toprule
\multicolumn{2}{l|}{{Method}}  & \multicolumn{2}{c|}{LOL-v2-real}  & \multicolumn{2}{c|}{LOL-v2-syn}& \multicolumn{2}{c}{Average}\\  
\multicolumn{2}{l|}{}  & \multicolumn{1}{c}{{PSNR}}  & \multicolumn{1}{c|}{{SSIM}}& \multicolumn{1}{c}{{PSNR}}  & \multicolumn{1}{c|}{{SSIM}}& \multicolumn{1}{c}{{PSNR}}  & \multicolumn{1}{c}{{SSIM}}\\  
\midrule

\multicolumn{2}{l|}{*EnGAN \cite{jiang2021enlightengan} {\scriptsize(TIP'21)}}  & \multicolumn{1}{c}{18.23} & \multicolumn{1}{c|}{0.617}& \multicolumn{1}{c}{16.57} & \multicolumn{1}{c|}{0.734}& \multicolumn{1}{c}{17.4} & \multicolumn{1}{c}{0.676}\\ 
\multicolumn{2}{l|}{*RUAS \cite{liu2021retinex} {\scriptsize(CVPR'21)}} & \multicolumn{1}{c}{18.37} & \multicolumn{1}{c|}{0.723} & \multicolumn{1}{c}{16.55} & \multicolumn{1}{c|}{0.652}& \multicolumn{1}{c}{17.46} & \multicolumn{1}{c}{0.688}\\
\multicolumn{2}{l|}{*QuadPrior~\cite{Wang_2024_CVPR} {\scriptsize(CVPR'24)}}  & \multicolumn{1}{c}{20.48} & \multicolumn{1}{c|}{0.811}& \multicolumn{1}{c}{16.11} & \multicolumn{1}{c|}{{0.758}}& \multicolumn{1}{c}{18.30} & \multicolumn{1}{c}{0.785}\\
\hline
\multicolumn{2}{l|}{KinD \cite{zhang2019kindling} {\scriptsize(MM'19)}}  & \multicolumn{1}{c}{14.74} & \multicolumn{1}{c|}{0.641}& \multicolumn{1}{c}{13.29} & \multicolumn{1}{c|}{0.578}& \multicolumn{1}{c}{14.02} & \multicolumn{1}{c}{0.610}\\ 
\multicolumn{2}{l|}{Uformer~\cite{wang2022uformer} {\scriptsize(CVPR'22)}} & \multicolumn{1}{c}{18.82} & \multicolumn{1}{c|}{0.771} & \multicolumn{1}{c}{19.66} & \multicolumn{1}{c|}{ 0.871}& \multicolumn{1}{c}{19.24} & \multicolumn{1}{c}{0.821}\\
\multicolumn{2}{l|}{Restormer \cite{zamir2022restormer} {\scriptsize(CVPR'22)}}  & \multicolumn{1}{c}{19.94} & \multicolumn{1}{c|}{0.827}& \multicolumn{1}{c}{21.41} & \multicolumn{1}{c|}{0.830}& \multicolumn{1}{c}{20.68} & \multicolumn{1}{c}{0.829}\\
\multicolumn{2}{l|}{MIRNet~\cite{zamir2020learning} {\scriptsize(ECCV'20)}} & \multicolumn{1}{c}{20.02} & \multicolumn{1}{c|}{0.820} & \multicolumn{1}{c}{21.94} & \multicolumn{1}{c|}{0.876}& \multicolumn{1}{c}{20.98} & \multicolumn{1}{c}{0.848}\\
\multicolumn{2}{l|}{Sparse~\cite{yang2021sparse} {\scriptsize(TIP'21)}} & \multicolumn{1}{c}{20.06} & \multicolumn{1}{c|}{0.816} & \multicolumn{1}{c}{22.05} & \multicolumn{1}{c|}{0.905}& \multicolumn{1}{c}{21.06} & \multicolumn{1}{c}{0.861}\\
\multicolumn{2}{l|}{MambaIR~\cite{guo2024mambair} {\scriptsize(ECCV'24)}} & \multicolumn{1}{c}{21.25} & \multicolumn{1}{c|}{0.831} & \multicolumn{1}{c}{25.55} & \multicolumn{1}{c|}{0.929}& \multicolumn{1}{c}{23.40} & \multicolumn{1}{c}{0.880}\\
\multicolumn{2}{l|}{SNR-Net~\cite{xu2022snr} {\scriptsize(CVPR'22)}} & \multicolumn{1}{c}{21.48} & \multicolumn{1}{c|}{\underline{0.849}} & \multicolumn{1}{c}{24.14} & \multicolumn{1}{c|}{0.928}& \multicolumn{1}{c}{22.81} & \multicolumn{1}{c}{{0.889}}\\
\multicolumn{2}{l|}{IGDFormer~\cite{WEN2025111033} {\scriptsize(PR'25)}} & \multicolumn{1}{c}{22.73} & \multicolumn{1}{c|}{{0.833}} & \multicolumn{1}{c}{25.33} & \multicolumn{1}{c|}{0.937}& \multicolumn{1}{c}{24.03} & \multicolumn{1}{c}{{0.885}}\\
\multicolumn{2}{l|}{Retinexformer~\cite{retinexformer} {\scriptsize(ICCV'23)}}  & \multicolumn{1}{c}{{22.80}} & \multicolumn{1}{c|}{{0.840}}& \multicolumn{1}{c}{{25.67}} & \multicolumn{1}{c|}{{0.930}}& \multicolumn{1}{c}{{24.24}} & \multicolumn{1}{c}{{0.885}}\\
\multicolumn{2}{l|}{MambaLLIE~\cite{weng2024mamballie} {\scriptsize(NeurIPS'24)}} & \multicolumn{1}{c}{\underline{22.95}} & \multicolumn{1}{c|}{0.847} & \multicolumn{1}{c}{\underline{25.87}} & \multicolumn{1}{c|}{\underline{0.940}}& \multicolumn{1}{c}{\underline{24.41}} & \multicolumn{1}{c}{\underline{0.894}}\\
\hline
\multicolumn{2}{l|}{HINT (Ours)} & \multicolumn{1}{c}{\textbf{23.11}} & \multicolumn{1}{c|}{\textbf{0.884}} & \multicolumn{1}{c}{\textbf{27.17}}  & \multicolumn{1}{c|}{\textbf{0.950}}& \multicolumn{1}{c}{\textbf{25.14}} & \multicolumn{1}{c}{\textbf{0.917}}\\ 
\bottomrule
\vspace{-5mm}
\end{tabular}}
\label{tab:enhancement_lolv2}
\end{table}

\noindent\textbf{Metrics.}
We employ the popular metrics, including peak signal-to-noise ratio (PSNR)~\cite{wang2004image} and structural similarity (SSIM), to evaluate the restored result with the reference image. 
Besides, the non-reference metric, i.e., MANIQA~\cite{yang2022maniqa}, is adopted to measure the restoration performance on real-world inputs. 
In the tables, we \textbf{highlight} and \underline{underline} the best and second-best scores, respectively. 

\begin{figure*}[htp!]
\footnotesize
\centering
\begin{tabular}{ccc}
\hspace{-0.42cm}
\begin{adjustbox}{valign=t}
\begin{tabular}{cccccccc}
\includegraphics[width=0.12\textwidth]{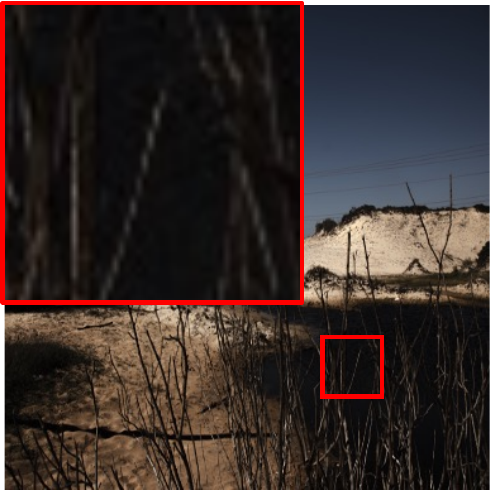} \hspace{-4mm} &
\includegraphics[width=0.12\textwidth]{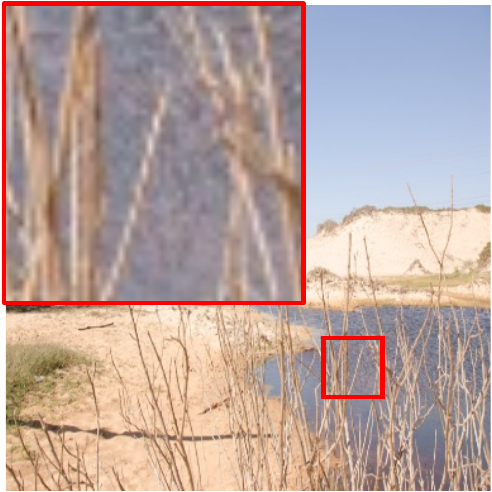} \hspace{-4mm} &
\includegraphics[width=0.12\textwidth]{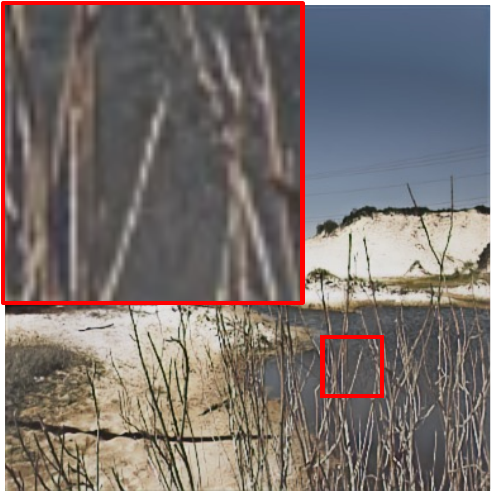} \hspace{-4mm} &
\includegraphics[width=0.12\textwidth]{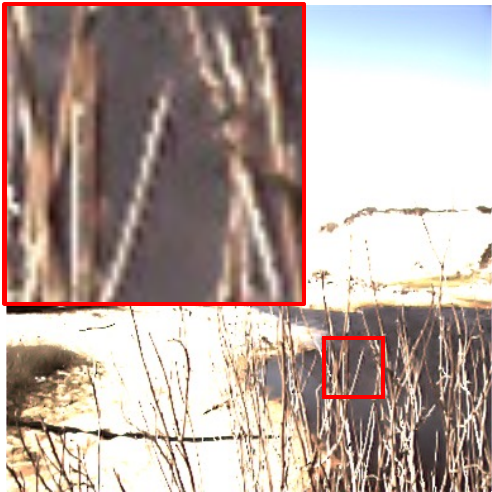} \hspace{-4mm} &
\includegraphics[width=0.12\textwidth]{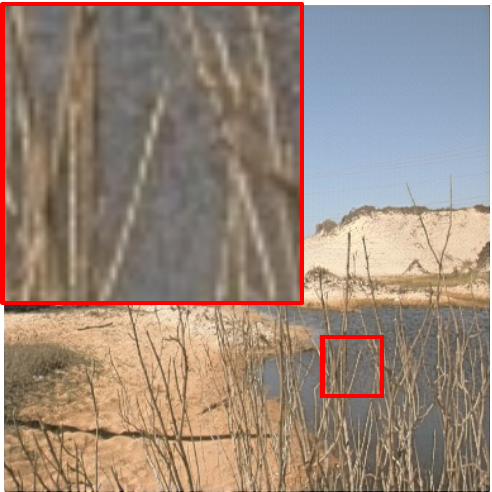} \hspace{-4mm} &
\includegraphics[width=0.12\textwidth]{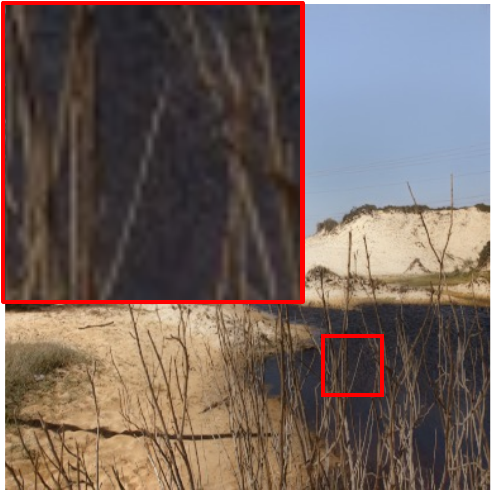} \hspace{-4mm} &
\includegraphics[width=0.12\textwidth]{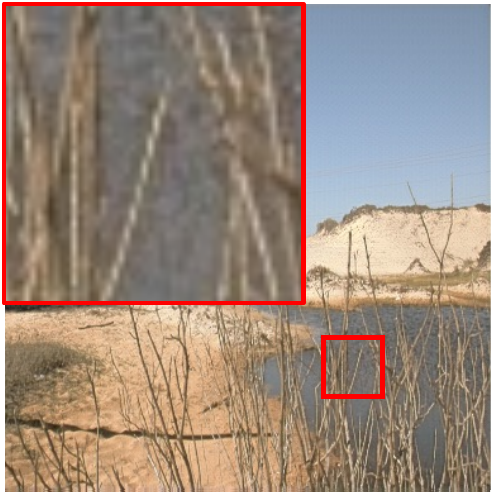} \hspace{-4mm} &
\includegraphics[width=0.12\textwidth]{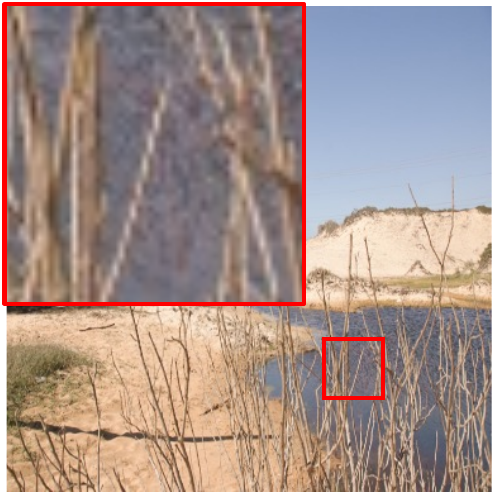} \hspace{-4mm} 
\\
\includegraphics[width=0.12\textwidth]{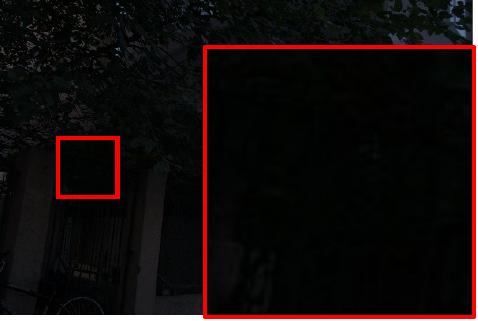} \hspace{-4mm} &
\includegraphics[width=0.12\textwidth]{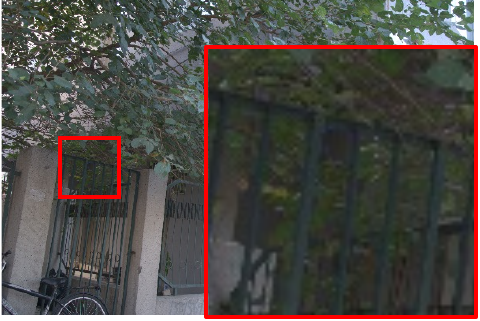} \hspace{-4mm} &
\includegraphics[width=0.12\textwidth]{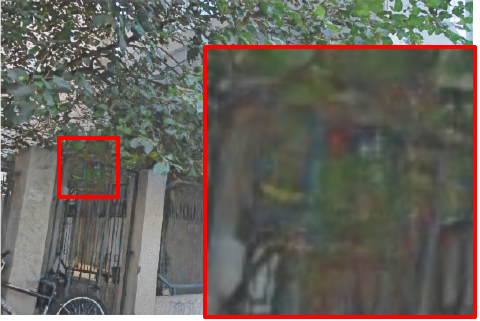} \hspace{-4mm} &
\includegraphics[width=0.12\textwidth]{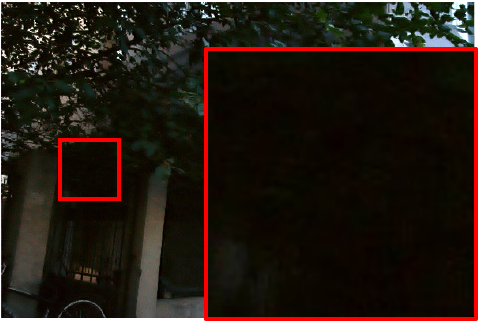} \hspace{-4mm} &
\includegraphics[width=0.12\textwidth]{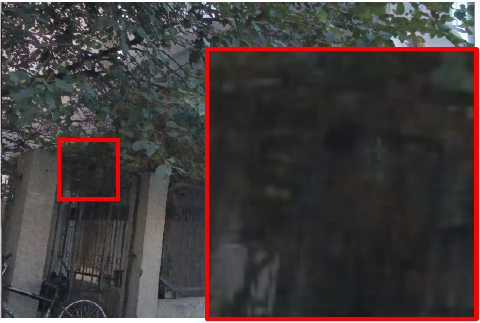} \hspace{-4mm} &
\includegraphics[width=0.12\textwidth]{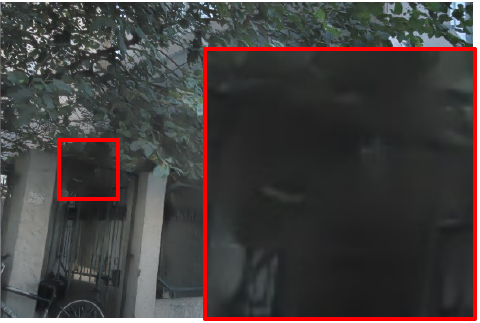} \hspace{-4mm} &
\includegraphics[width=0.12\textwidth]{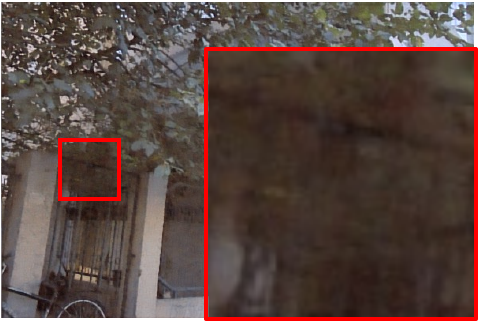} \hspace{-4mm} &
\includegraphics[width=0.12\textwidth]{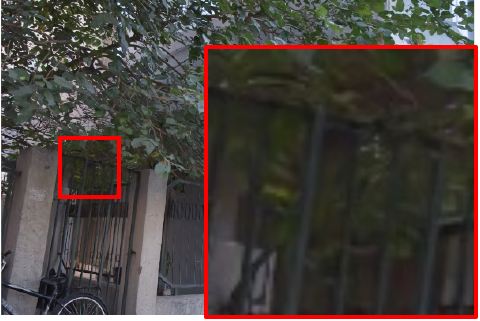} \hspace{-4mm} 
\\
(a) Input \hspace{-4mm} &
(b) Reference \hspace{-4mm} &
(c) KinD~\cite{zhang2019kindling} \hspace{-4mm} &
(d) RUAS~\cite{liu2021retinex} \hspace{-4mm} &
(e) MIRNet~\cite{zamir2020learning} \hspace{-4mm} &
(f) {Sparse}~\cite{yang2021sparse} \hspace{-4mm} &
(g) {Restormer}~\cite{zamir2022restormer} \hspace{-4mm} &
(h) {HINT} (Ours) \hspace{-4mm} 

\\
\end{tabular}
\end{adjustbox}
\end{tabular}
\vspace{-3mm}
\caption{Qualitative results on LOL-v2~\cite{yang2021sparse} for low-light enhancement. 
The top case is from the synthetic subset, whereas the bottom one is from the real subset. 
Compared to other techniques, HINT generates vivid images without introducing noticeable color distortion. 
}
\vspace{-4mm}
\label{pic:enhancement_lolv2}
\end{figure*}

\begin{figure*}[htp!]
\footnotesize
\centering
\begin{tabular}{ccc}
\hspace{-0.42cm}
\begin{adjustbox}{valign=t}
\begin{tabular}{ccccccc}
\includegraphics[width=0.135\textwidth]{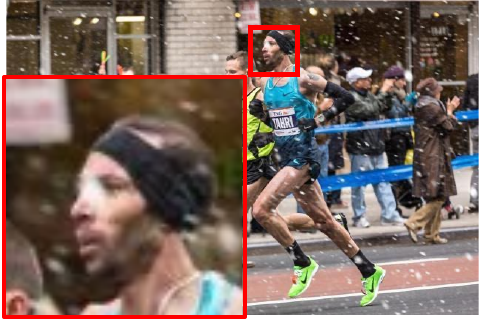} \hspace{-4mm} &
\includegraphics[width=0.135\textwidth]{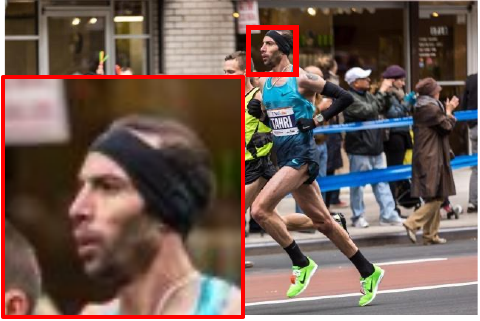} \hspace{-4mm} &
\includegraphics[width=0.135\textwidth]{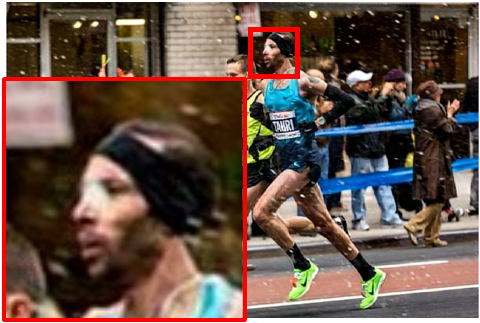} \hspace{-4mm} &
\includegraphics[width=0.135\textwidth]{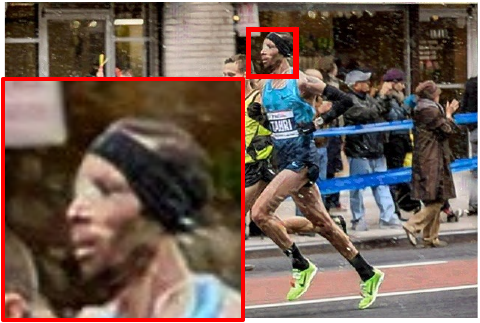} \hspace{-4mm} &
\includegraphics[width=0.135\textwidth]{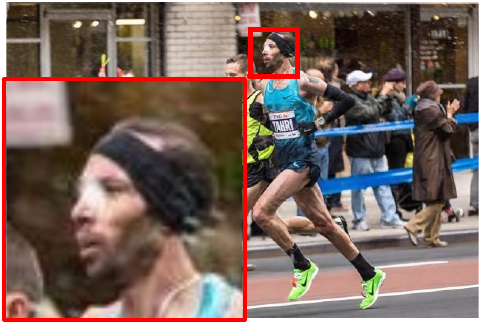} \hspace{-4mm} &
\includegraphics[width=0.135\textwidth]{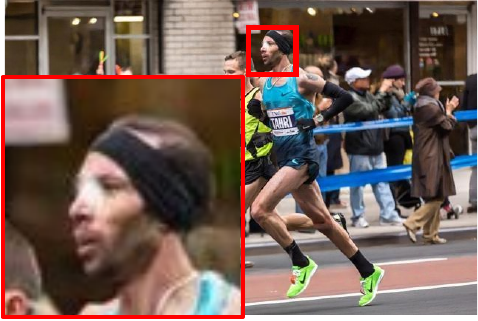} \hspace{-4mm} &
\includegraphics[width=0.135\textwidth]{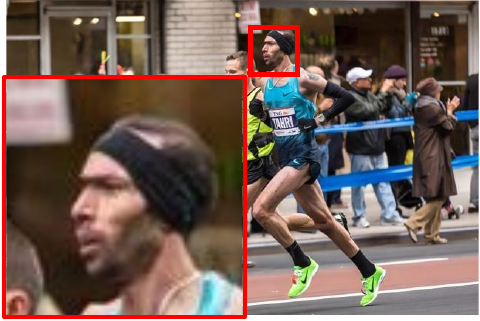} \hspace{-4mm} 
\\
(a) Input \hspace{-4mm} &
(b) Reference \hspace{-4mm} &
(c) HDCW~\cite{Chen_2021_ICCV} \hspace{-4mm} &
(d) JSTASR~\cite{chen2020jstasr} \hspace{-4mm} &
(e) Uformer~\cite{wang2022uformer} \hspace{-4mm} &
(f) {AST}~\cite{zhou2024AST} \hspace{-4mm} &
(g) {HINT} (Ours) \hspace{-4mm} 
\\
\end{tabular}
\end{adjustbox}
\end{tabular}
\vspace{-3mm}
\caption{Qualitative results on Snow100K~\cite{liu2018desnownet} for snow removal. 
HINT offers a clear result, while the images generated by other considered approaches remain noticeable snow artifacts.
}
\vspace{-5mm}
\label{pic:vis_snow}
\end{figure*}

\subsection{Main Results}
\noindent\textbf{Low-Light Enhancement.}
We compare the proposed HINT with state-of-the-art low-light enhancement methods on the LOL-v2 dataset in Table~\ref{tab:enhancement_lolv2}. 
As one can see, HINT performs favorably against considered approaches in terms of PSNR and SSIM on both real and synthetic subsets. 
In particular, when averaged across both subsets, HINT achieves a significant 0.9 dB improvement on PSNR over Retinexformer~\cite{retinexformer}, which is the first Transformer-based pipeline designed specifically for low-light enhancement. 
Compared to general image restoration schemes based on diverse architectures, e.g., CNN-based~\cite{zamir2020learning}, Transformer-based~\cite{wang2022uformer,zamir2022restormer}, and Mamba-based~\cite{guo2024mambair}, HINT receives at least 1.74 dB gain. 
Notably, our model yields a remarkable 1.11 dB performance improvement on PSNR over the recent algorithm IGDFormer~\cite{WEN2025111033}. 
The qualitative comparison is depicted in Figure~\ref{pic:enhancement_lolv2}. 
In the top row, previous methods fall short of recovering satisfactory results. 
They either meet over-/under-exposed issue~\cite{liu2021retinex,yang2021sparse} (Figure~\ref{pic:enhancement_lolv2}d and \ref{pic:enhancement_lolv2}f) or struggle to restore the true colors~\cite{zhang2019kindling,zamir2022restormer,zamir2020learning} (Figure~\ref{pic:enhancement_lolv2}c, \ref{pic:enhancement_lolv2}e and \ref{pic:enhancement_lolv2}g). 
For the bottom case, the image restored by HINT is closer to the reference one, where other algorithms introduce noticeable blur patterns~\cite{zhang2019kindling,zamir2020learning,yang2021sparse,zamir2022restormer} (Figure~\ref{pic:enhancement_lolv2}c, \ref{pic:enhancement_lolv2}e, \ref{pic:enhancement_lolv2}f, \ref{pic:enhancement_lolv2}g), or cause a under-exposed problem~\cite{liu2021retinex} (Figure~\ref{pic:enhancement_lolv2}d).

\noindent\textbf{Snow Removal.}
For image desnowing, we conduct experiments on the Snow100K~\cite{liu2018desnownet} dataset. 
HINT archives the best PSNR score and a competitive SSIM result among all methods. 
To be specific, HINT provides a significant 1.64 dB boost on PSNR over the recent general restoration pipeline AST~\cite{zhou2024AST}. 
Additionally, compared to approaches~\cite{chen2020jstasr,liu2018desnownet,Chen_2021_ICCV} that are designed for desnowing, HINT showcases the superiority of better performance. 
Furthermore, compared to methods~\cite{Li_2020_CVPR,Valanarasu_2022_CVPR} that are proposed to address adverse weather conditions, our method earns consistent benefits. 
For general restoration schemes, including CNN-based~\cite{chen2022simple,cui2023focal,cui2023selective,convIr_pami24} and Transformer-based~\cite{wang2022uformer,zhou2024AST}, HINT shows at least a 0.35 dB performance boost on PSNR.
The visual comparisons are shown in Figure~\ref{pic:vis_snow}, where HINT restores a clear result (Figure~\ref{pic:vis_snow}g). 
The desnowing techniques~\cite{Chen_2021_ICCV,chen2020jstasr} offer unsatisfactory outputs (Figure~\ref{pic:vis_snow}c and \ref{pic:vis_snow}d). 
The results from Transformer-based approaches~\cite{zhou2024AST,wang2022uformer} exhibit noticeable snow artifacts (Figure~\ref{pic:vis_snow}e and \ref{pic:vis_snow}f). 

\begin{table}\footnotesize
\caption{Quantitative results on Snow100K \cite{liu2018desnownet} for snow removal. }
\vspace{-2mm}
\centering
\footnotesize
\setlength{\tabcolsep}{.1305mm}
{
\begin{tabular}{ccccccccccccccc}
\toprule[0.8pt]
\multicolumn{1}{l}{\multirow{2}{*}{Method}} & \multicolumn{1}{c}{JSTASR}   & \multicolumn{1}{c}{All in One} 
& \multicolumn{1}{c}{Uformer} &\multicolumn{1}{c}{DesnowNet}
&\multicolumn{1}{c}{HDCW}&\multicolumn{1}{c}{TransWeather}\\ 
\multicolumn{1}{l}{} & \multicolumn{1}{c}{{\scriptsize ECCV'20}}   & \multicolumn{1}{c}{{\scriptsize CVPR'20}} 
& \multicolumn{1}{c}{{\scriptsize CVPR'22}} &\multicolumn{1}{c}{{\scriptsize TIP'18}}
&\multicolumn{1}{c}{{\scriptsize ICCV'21}}&\multicolumn{1}{c}{{\scriptsize CVPR'22}}\\ 
\multicolumn{1}{l}{{}} & \multicolumn{1}{c}{\cite{chen2020jstasr}}   & \multicolumn{1}{c}{\cite{Li_2020_CVPR}} 
& \multicolumn{1}{c}{\cite{wang2022uformer}} &\multicolumn{1}{c}{\cite{liu2018desnownet}}
&\multicolumn{1}{c}{\cite{Chen_2021_ICCV}}&\multicolumn{1}{c}{\cite{Valanarasu_2022_CVPR}}\\ 
\hline
\multicolumn{1}{l}{{PSNR}} & \multicolumn{1}{c}{23.12}   & \multicolumn{1}{c}{26.07} 
& \multicolumn{1}{c}{29.80 } &\multicolumn{1}{c}{30.50 }
&\multicolumn{1}{c}{31.54}&\multicolumn{1}{c}{31.82} 
\\
\multicolumn{1}{l}{{SSIM}} & \multicolumn{1}{c}{0.86}   & \multicolumn{1}{c}{0.88} 
& \multicolumn{1}{c}{0.93} &\multicolumn{1}{c}{0.94}
&\multicolumn{1}{c}{\underline{0.95}}&\multicolumn{1}{c}{0.93} 
\\
\midrule[0.8pt]
\multicolumn{1}{l}{\multirow{2}{*}{Method}} 
& \multicolumn{1}{c}{NAFNet} & \multicolumn{1}{c}{AST} 
&\multicolumn{1}{c}{FocalNet}&\multicolumn{1}{c}{SFNet}&\multicolumn{1}{c}{ConvIR-S}&\multicolumn{1}{c}{HINT}\\ 
\multicolumn{1}{l}{} 
& \multicolumn{1}{c}{{\scriptsize ECCV'22}} & \multicolumn{1}{c}{{\scriptsize CVPR'24}} 
&\multicolumn{1}{c}{{\scriptsize ICCV'23}}&\multicolumn{1}{c}{{\scriptsize ICLR'23}}&\multicolumn{1}{c}{{\scriptsize TPAMI'24}}&\multicolumn{1}{c}{{\scriptsize -}}\\ 
\multicolumn{1}{l}{} 
& \multicolumn{1}{c}{\cite{chen2022simple}} & \multicolumn{1}{c}{\cite{zhou2024AST}} 
&\multicolumn{1}{c}{\cite{cui2023focal}}&\multicolumn{1}{c}{\cite{cui2023selective}}&\multicolumn{1}{c}{\cite{convIr_pami24}}&\multicolumn{1}{c}{(Ours)}\\ 
\hline
\multicolumn{1}{l}{{PSNR}} 
& \multicolumn{1}{c}{32.41} & \multicolumn{1}{c}{32.50} 
&\multicolumn{1}{c}{33.53} &\multicolumn{1}{c}{33.79}&\multicolumn{1}{c}{33.79}&\multicolumn{1}{c}{\textbf{34.14}} 
\\
\multicolumn{1}{l}{{SSIM}} 
& \multicolumn{1}{c}{\underline{0.95}} & \multicolumn{1}{c}{\textbf{0.96}} 
&\multicolumn{1}{c}{\underline{0.95}}&\multicolumn{1}{c}{\underline{0.95}}&\multicolumn{1}{c}{\underline{0.95}}&\multicolumn{1}{c}{0.94} 
\\
\bottomrule[0.8pt]
\vspace{-5mm}
\end{tabular}}
\label{table4desnow_appdix}
\end{table}

\begin{table}\footnotesize
\caption{Quantitative comparison on SOTS~\cite{li2018benchmarking} for haze removal.}
\vspace{-2mm}
\centering
\scriptsize
\setlength{\tabcolsep}{.15mm}
{
\begin{tabular}{cccccccccccc}
\toprule[0.8pt]
\multicolumn{1}{l}{\multirow{2}{*}{Method}} & 
\multicolumn{1}{c}{EPDN}& \multicolumn{1}{c}{FDGAN~} & \multicolumn{1}{c}{AirNet}&  \multicolumn{1}{c}{InstructIR}&  \multicolumn{1}{c}{Restormer}& 
\multicolumn{1}{c}{NAFNet} \\
\multicolumn{1}{l}{} & 
\multicolumn{1}{c}{{\scriptsize CVPR'19}}& \multicolumn{1}{c}{{\scriptsize AAAI'20}} & \multicolumn{1}{c}{{\scriptsize CVPR'22}}&  \multicolumn{1}{c}{{\scriptsize ECCV'24}}&  \multicolumn{1}{c}{{\scriptsize CVPR'22}}&  \multicolumn{1}{c}{{\scriptsize ECCV'22}}\\
\multicolumn{1}{l}{} & 
\multicolumn{1}{c}{\cite{qu2019enhanced}}& \multicolumn{1}{c}{\cite{dong2020fd}} & \multicolumn{1}{c}{\cite{li2022all}}&  \multicolumn{1}{c}{\cite{conde2024high}}&  \multicolumn{1}{c}{\cite{zamir2022restormer}}&  \multicolumn{1}{c}{\cite{chen2022simple}}\\
\hline
\multicolumn{1}{l}{{PSNR}} 
&\multicolumn{1}{c}{22.57}& \multicolumn{1}{c}{ 23.15} & \multicolumn{1}{c}{23.18} & \multicolumn{1}{c}{ 30.22 } 
& \multicolumn{1}{c}{ 30.87}& \multicolumn{1}{c}{ 32.41} \\ 
\multicolumn{1}{l}{{SSIM}} 
& \multicolumn{1}{c}{0.863}& \multicolumn{1}{c}{0.921} & \multicolumn{1}{c}{ 0.900}&  \multicolumn{1}{c}{{0.959}}&  \multicolumn{1}{c}{{0.969}}&  \multicolumn{1}{c}{{0.970}}\\ 
\midrule[0.5pt]
\multicolumn{1}{l}{\multirow{2}{*}{Method}} 
&\multicolumn{1}{c}{FSNet}& \multicolumn{1}{c}{PromptIR} &\multicolumn{1}{c}{DehazeFormer}& \multicolumn{1}{c}{AdaIR} & \multicolumn{1}{c}{NDR-Restore}   &\multicolumn{1}{c}{HINT}\\ 
\multicolumn{1}{l}{} & 
\multicolumn{1}{c}{{\scriptsize TPAMI'24}}& \multicolumn{1}{c}{{\scriptsize NeurIPS'23}} & \multicolumn{1}{c}{{\scriptsize TIP'23}}&  \multicolumn{1}{c}{{\scriptsize ICLR'25}}& \multicolumn{1}{c}{{\scriptsize TIP'24}}& \multicolumn{1}{c}{-}\\
\multicolumn{1}{l}{} 
&\multicolumn{1}{c}{\cite{FSNet_pami24}}& \multicolumn{1}{c}{\cite{potlapalli2024promptir}} &\multicolumn{1}{c}{\cite{10076399}}& \multicolumn{1}{c}{\cite{cui2025adair}} & \multicolumn{1}{c}{\cite{NDR-Restore_tip24}}  &\multicolumn{1}{c}{(Ours)}\\ 
\hline
\multicolumn{1}{l}{{PSNR}}
&\multicolumn{1}{c}{31.11}& \multicolumn{1}{c}{31.31}  &\multicolumn{1}{c}{31.78} &\multicolumn{1}{c}{31.80} &\multicolumn{1}{c}{\underline{31.96}} &\multicolumn{1}{c}{\textbf{32.24}}\\ 
\multicolumn{1}{l}{{SSIM}} 
&\multicolumn{1}{c}{0.971}& \multicolumn{1}{c}{0.973} &\multicolumn{1}{c}{{0.977}}&\multicolumn{1}{c}{\textbf{0.981}} & \multicolumn{1}{c}{\underline{0.980}} &\multicolumn{1}{c}{\textbf{0.981}}\\ 
\bottomrule[0.8pt]
\vspace{-5mm}
\end{tabular}}
\label{tab:deSynHaze_SOT}
\end{table}

\begin{figure}
\scriptsize
\centering
\begin{tabular}{ccc}
\begin{adjustbox}{valign=t}
\begin{tabular}{ccccc}
\hspace{-3mm}\includegraphics[width=0.088\textwidth]{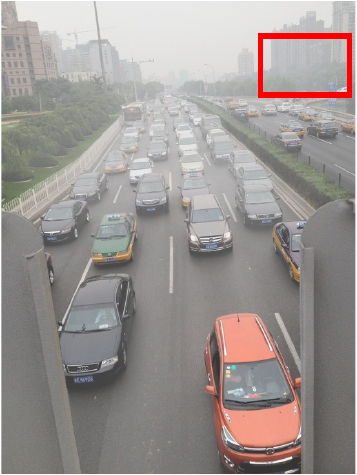}  &
\hspace{-3mm}\includegraphics[width=0.088\textwidth]{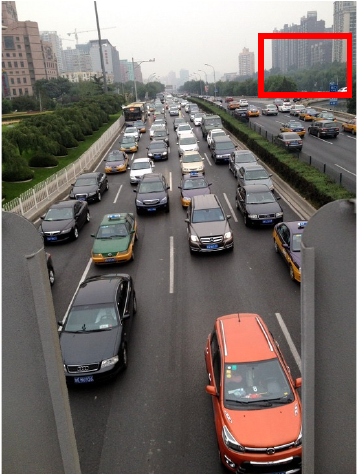}   &
\hspace{-3mm}\includegraphics[width=0.088\textwidth]{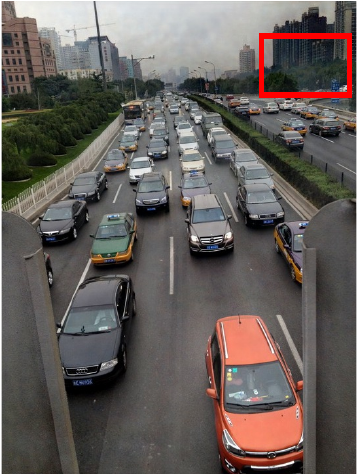}  &
\hspace{-3mm}\includegraphics[width=0.088\textwidth]{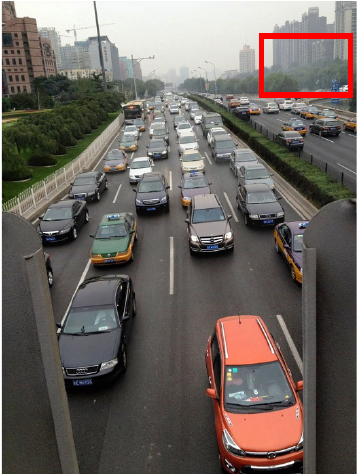}  &
\hspace{-3mm}\includegraphics[width=0.088\textwidth]{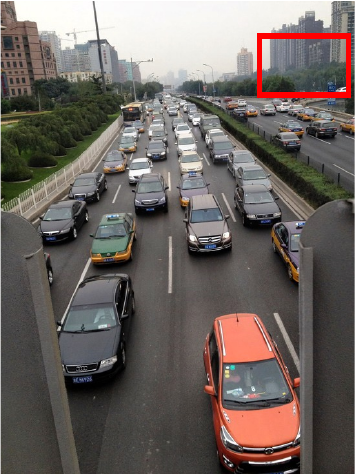}  
\\
\hspace{-3mm}\includegraphics[width=0.088\textwidth]{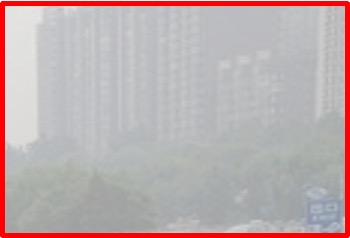}  &
\hspace{-3mm}\includegraphics[width=0.088\textwidth]{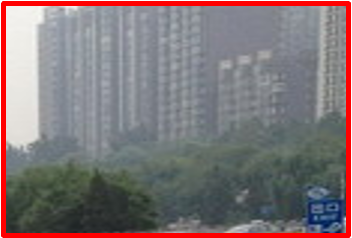}   &
\hspace{-3mm}\includegraphics[width=0.088\textwidth]{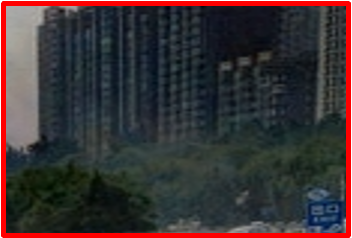}  &
\hspace{-3mm}\includegraphics[width=0.088\textwidth]{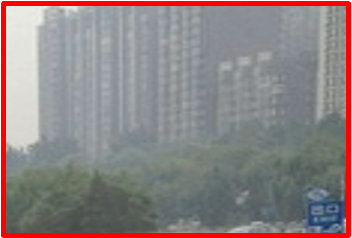}  &
\hspace{-3mm}\includegraphics[width=0.088\textwidth]{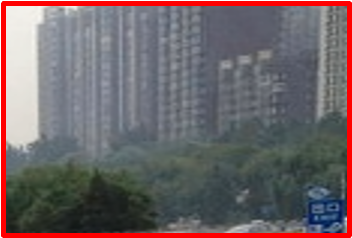}  
\\
\hspace{-3mm}(a) &
\hspace{-3mm}(b)   &   
\hspace{-3mm}(c)   &
\hspace{-3mm}(d)   &
\hspace{-3mm}(e) 
\\
\hspace{-3mm} Input &
\hspace{-3mm} Reference  &   
\hspace{-3mm} AirNet \cite{li2022all}  &
\hspace{-3mm} PromptIR \cite{potlapalli2024promptir}  &
\hspace{-3mm} HINT
\\
\end{tabular}
\end{adjustbox}
\end{tabular}
\vspace{-2mm}
\caption{
Qualitative results on SOTS~\cite{li2018benchmarking} benchmark for haze removal. 
The image generated by HINT is closer to the reference one, compared to other methods.
}
\vspace{-5mm}
\label{pic:vis_dehaze_sots}
\end{figure}

\begin{figure*}[htp!]
\footnotesize
\centering
\begin{tabular}{ccc}
\hspace{-0.42cm}
\begin{adjustbox}{valign=t}
\begin{tabular}{ccccccc}
\includegraphics[width=0.135\textwidth]{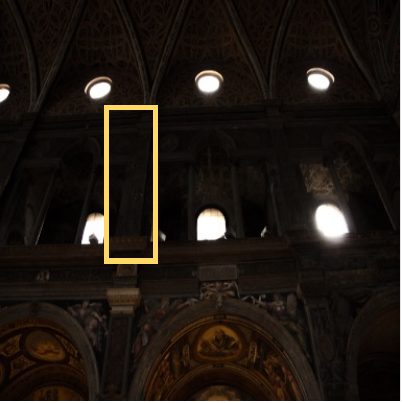} \hspace{-4mm} &
\includegraphics[width=0.135\textwidth]{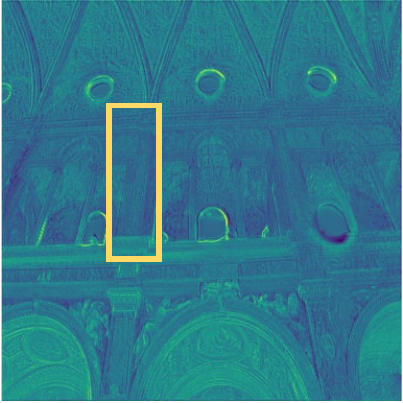} \hspace{-4mm} &
\includegraphics[width=0.135\textwidth]{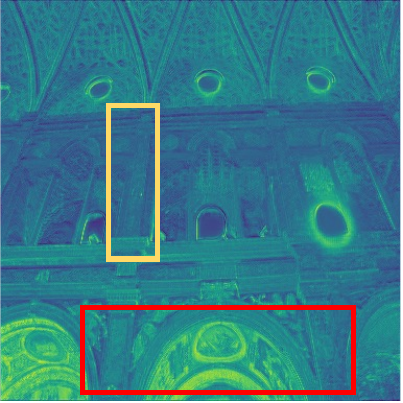} \hspace{-4mm} &
\includegraphics[width=0.135\textwidth]{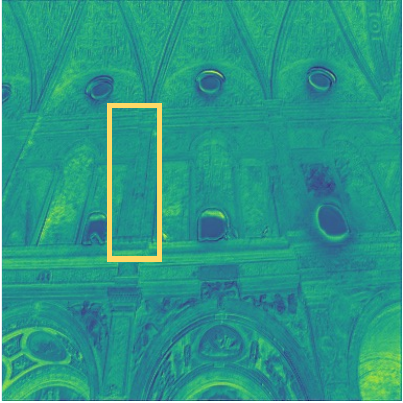} \hspace{-4mm} &
\includegraphics[width=0.135\textwidth]{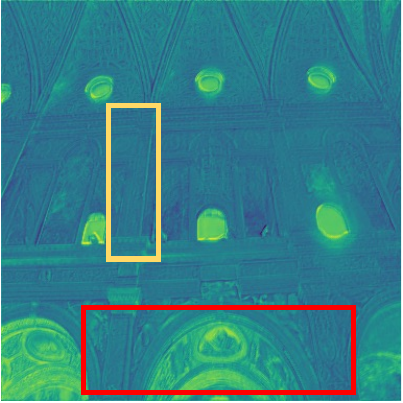} \hspace{-4mm} &
\includegraphics[width=0.135\textwidth]{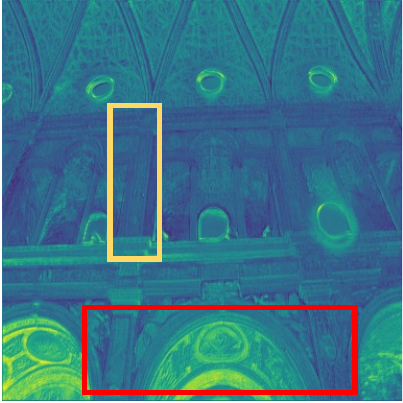} \hspace{-4mm} &
\includegraphics[width=0.135\textwidth]{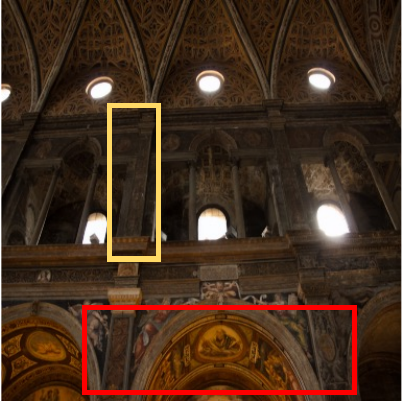} \hspace{-4mm} 
\\
(a) Input \hspace{-4mm} &
(b) head1 in MDTA \hspace{-4mm} &
(c) head2 in MDTA \hspace{-4mm} &
(d) head3 in MDTA \hspace{-4mm} &
(e) head4 in MDTA \hspace{-4mm} &
(f) avg heads in MDTA \hspace{-4mm} &
(g) res of MDTA \hspace{-4mm} 
\\
\includegraphics[width=0.135\textwidth]{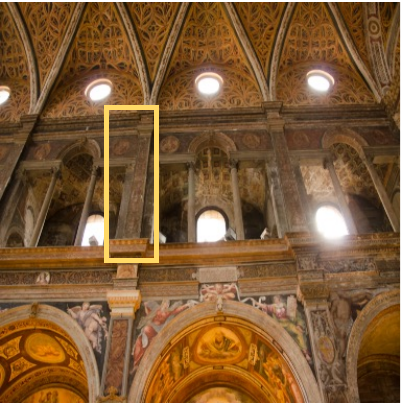} \hspace{-4mm} &
\includegraphics[width=0.135\textwidth]{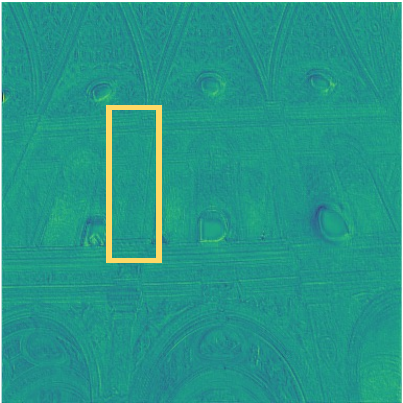} \hspace{-4mm} &
\includegraphics[width=0.135\textwidth]{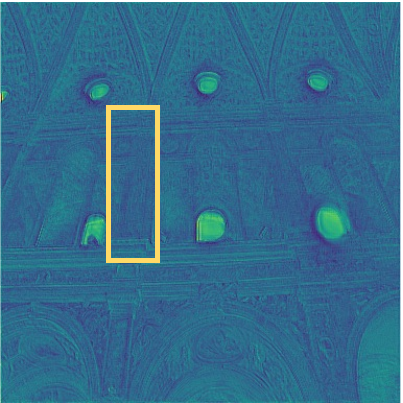} \hspace{-4mm} &
\includegraphics[width=0.135\textwidth]{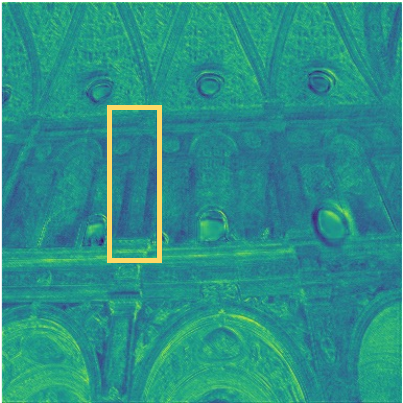} \hspace{-4mm} &
\includegraphics[width=0.135\textwidth]{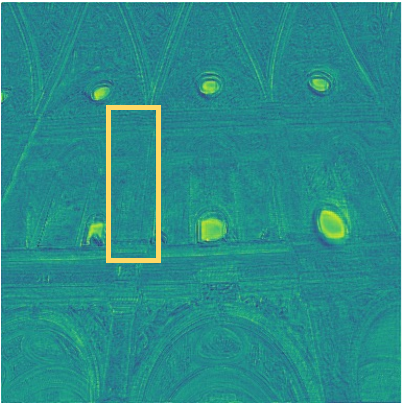} \hspace{-4mm} &
\includegraphics[width=0.135\textwidth]{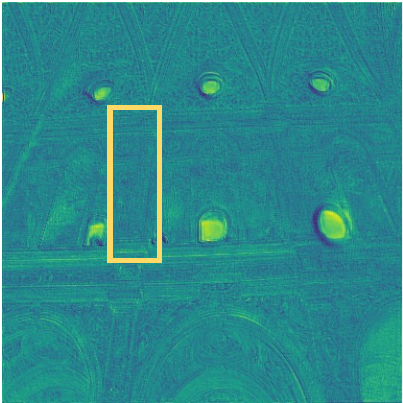} \hspace{-4mm} &
\includegraphics[width=0.135\textwidth]{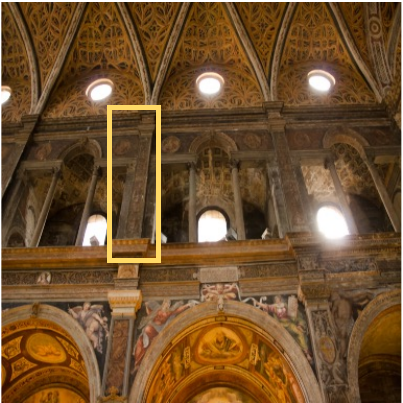} \hspace{-4mm} 
\\
(a) Reference \hspace{-4mm} &
(b) head1 in HMHA \hspace{-4mm} &
(c) head2 in HMHA \hspace{-4mm} &
(d) head3 in HMHA \hspace{-4mm} &
(e) head4 in HMHA \hspace{-4mm} &
(f) avg heads in HMHA \hspace{-4mm} &
(g) res of HMHA \hspace{-4mm} 

\\
\end{tabular}
\end{adjustbox}
\end{tabular}
\vspace{-2mm}
\caption{Feature Visualization. 
The top line exhibits the feature map of each head learned by MDTA, while the feature maps at the bottom illustrate the results of HMHA. 
The heads in MDTA intend to focus on the same regions (\textcolor{red}{red} boxes), whereas the counterparts in HMHA showcase superiority in learning representations from different subspaces. 
As a result, the model equipped with the proposed HMHA restores a pleasant image, which is closer to the reference one (\textcolor{yellow}{yellow} boxes).
}
\vspace{-4mm}
\label{pic:ablation_vis_MDTA_HMSA}
\end{figure*}

\noindent\textbf{Haze Removal.}
We perform image dehazing experiments on the SOTS~\cite{li2018benchmarking} dataset, where 11 representative approaches are considered to make a comparison in Table~\ref{tab:deSynHaze_SOT}. 
As can be seen, HINT outperforms all other methods on both PSNR and SSIM metrics. 
It is worth noting that, HINT surpasses all the methods~\cite{qu2019enhanced,dong2020fd,10076399}, which are elaborately designed for dehazing, by at least 0.46 dB on PSNR. 
When compared to image restoration pipelines (all-in-one paradigms~\cite{cui2025adair,potlapalli2024promptir,NDR-Restore_tip24} and general ones~\cite{conde2024high,zamir2022restormer,chen2022simple,FSNet_pami24}), HINT consistently shows advantages. 
Figure~\ref{pic:vis_dehaze_sots} illustrates the qualitative results. 
In comparison, other methods introduce either color distortion issue~\cite{li2022all} (Figure~\ref{pic:vis_dehaze_sots}c) or haze effect~\cite{potlapalli2024promptir} (Figure~\ref{pic:vis_dehaze_sots}d), while HINT restores a vivid result. 

\subsection{Analysis and Discussion}
\label{subsec:ablation}
We have demonstrated that exploring the HMHA mechanism equipped with QKCU in a Transformer-based model brings benefits across several restoration tasks. 
%
Next, we further perform experiments to study the proposed modules and analyze the effect of each component. 
For ablation studies, various low-light enhancement models HINT are trained on the LOL-v2-syn subset~\cite{yang2021sparse}. 
To ensure fair comparisons, all models are trained under identical experimental settings, and the FLOPs/Runtimes are computed using 256 × 256 input images.

\begin{table}\footnotesize
\caption{Ablation study for different self-attention mechanisms. }
\vspace{-2mm}
\centering
\footnotesize

\setlength{\tabcolsep}{2.65mm}
{
\begin{tabular}{cccccccc}
\toprule[0.8pt]
\multicolumn{2}{l|}{\multirow{2}{*}{Model}} & \multicolumn{2}{c}{W-MSA}& \multicolumn{2}{c}{MDTA} &\multicolumn{2}{|c}{HMHA} \\ 
\multicolumn{2}{l|}{}   & \multicolumn{2}{c}{\cite{wang2022uformer}} & \multicolumn{2}{c}{\cite{zamir2022restormer}} &\multicolumn{2}{|c}{Ours} \\ 
\midrule[0.8pt]
\multicolumn{2}{l|}{PSNR/SSIM} 
&\multicolumn{2}{c}{24.19/0.941}   
& \multicolumn{2}{c}{26.42/0.948} 
& \multicolumn{2}{|c}{27.17/0.950}    \\
\bottomrule[0.8pt]
\end{tabular}}
\label{tab:abs_sa}
\vspace{-3mm}
\end{table}
\begin{table}
\caption{Ablation study of HMHA. 
}
\vspace{-2.mm}
\centering
\footnotesize
\setlength{\tabcolsep}{3.05mm}
{
\begin{tabular}{ccccc}
\toprule[0.8pt]
\multicolumn{1}{c}{}
& \multicolumn{1}{c}{{Ranking Strategy}}
& \multicolumn{1}{|c}{{Params}}& \multicolumn{1}{c}{{PSNR}}
& \multicolumn{1}{c}{{SSIM}} \\ \midrule[0.8pt]

\multicolumn{1}{l}{(a)} &\multicolumn{1}{c}{No-Ranking~\cite{zamir2022restormer}} & \multicolumn{1}{|c}{{24.76}}&\multicolumn{1}{c}{26.42} &  \multicolumn{1}{c}{0.948} \\
  
\multicolumn{1}{l}{(b)} &\multicolumn{1}{c}{Random Shuffle~\cite{zhang2018shufflenet}} & \multicolumn{1}{|c}{{24.87}}&\multicolumn{1}{c}{26.54} &  \multicolumn{1}{c}{0.949} \\
  
\multicolumn{1}{l}{(c)} &\multicolumn{1}{c}{HMHA (Ours)} & \multicolumn{1}{|c}{{24.87}}&\multicolumn{1}{c}{27.17} &  \multicolumn{1}{c}{0.950}  \\

\bottomrule
\end{tabular}}\textbf{}

\label{tab:ablation-similarity}
\vspace{-3mm}
\end{table}

\begin{table}
\caption{Ablation study of QKCU. 
}
\vspace{-2.mm}
\centering
\footnotesize
\setlength{\tabcolsep}{2.5mm}
{
\begin{tabular}{ccccccccccccc}
\toprule[0.8pt]
\multicolumn{1}{c}{} &\multicolumn{2}{c}{IntraCache} & \multicolumn{2}{c}{{InterCache}}
& \multicolumn{2}{|c}{{Params}}
& \multicolumn{2}{c}{{PSNR}}&  \multicolumn{2}{c}{{SSIM}}  \\ \midrule[0.8pt]

\multicolumn{1}{c}{(a)} &\multicolumn{2}{c}{}    & \multicolumn{2}{c}{}
  &\multicolumn{2}{|c}{21.34} &  \multicolumn{2}{c}{26.47} &  \multicolumn{2}{c}{0.949}\\

\multicolumn{1}{c}{(b)} &\multicolumn{2}{c}{$\surd$}    & \multicolumn{2}{c}{}&
\multicolumn{2}{|c}{23.82} &  \multicolumn{2}{c}{26.67} &  \multicolumn{2}{c}{0.949} \\

\multicolumn{1}{c}{(c)} &\multicolumn{2}{c}{}    & \multicolumn{2}{c}{$\surd$}&
\multicolumn{2}{|c}{22.39} &  \multicolumn{2}{c}{26.72} &  \multicolumn{2}{c}{0.949}\\
\multicolumn{1}{c}{(d)} &\multicolumn{2}{c}{$\surd$}    & \multicolumn{2}{c}{$\surd$}& \multicolumn{2}{|c}{24.87} &  \multicolumn{2}{c}{27.17} &  \multicolumn{2}{c}{0.950} \\
\bottomrule
\end{tabular}}
\label{tab:ablation-component}
\vspace{-3.mm}
\end{table}

\noindent\textbf{Effect of HMHA.}
To investigate whether the proposed HMHA enhances the capability of modeling diverse contextual information, we compare it with two representative variants, including (1) Window-based Multi-head Self-Attention (W-MSA)~\cite{liu2022tuformer}, and (2) Multi-Dconv Head Transposed Attention (MDTA)~\cite{zamir2022restormer}. 
The quantitative comparison is reported in Table~\ref{tab:abs_sa}. 
The model incorporating HMHA achieves the highest scores on both PSNR and SSIM metrics, outperforming other variants. 
Specifically, replacing HMHA with W-MSA or MDTA leads to significant performance drops of 2.98 dB and 0.75 dB, respectively. 
To investigate the effect of HMHA, we further visualize the feature maps learned by the heads in MDTA and HMHA, respectively. 
As shown in Figure~\ref{pic:ablation_vis_MDTA_HMSA}, the heads in MDTA intend to focus on the same regions (indicated by red boxes), whereas the heads in HMHA showcase superiority in learning distinct representations from different subspaces. 
Consequently, the model with HMHA produces a result that is closer to the reference image (highlighted by yellow boxes). 

We then show the effectiveness of the reranking strategy in HMHA for achieving discriminative representation in Table~\ref{tab:ablation-similarity}. 
When we replace the reranking strategy (Table~\ref{tab:ablation-similarity}c) with a random shuffle operation~\cite{zhang2018shufflenet} (Table~\ref{tab:ablation-similarity}b), we observe a performance drop of 0.63 dB on PSNR. 
In contrast, compared to the model without any ranking mechanism (Table~\ref{tab:ablation-similarity}a), which is degraded to vanilla MHA~\cite{zamir2022restormer}, our HMHA achieves a significant 0.75 performance boost. 
These results highlight the positive impact of the reranking strategy in enhancing the expressive power of the model. 

\noindent\textbf{Effect of QKCU.} 
To show the effectiveness of the proposed QKCU module, we conduct experiments on different variants in Table~\ref{tab:ablation-component}. 
Disabling either IntraCache or InterCache within QKCU results in performance declines of 0.45 dB and 0.5 dB, respectively, indicating the importance of these modules in restoring high-quality images. 
In total, the QKCU mechanism brings 0.7 dB performance gain with limited computation loads (16.5$\%$ in Parameters).

\begin{table}[t]
\caption{Model efficiency analysis on LOL-v2-syn~\cite{wang2019spatial}.}
\vspace{-2mm}
\centering
\footnotesize
\setlength{\tabcolsep}{.4mm}
{
\begin{tabular}{ccccccc}
\toprule[0.8pt]
\multicolumn{1}{l|}{{Method}} & \multicolumn{1}{c}{IPT~\cite{chen2021pre}}& \multicolumn{1}{c}{MIRNet~\cite{zamir2020learning}} & \multicolumn{1}{c}{Uformer~\cite{wang2022uformer}} & \multicolumn{1}{c|}{Restormer~\cite{zamir2022restormer}}& \multicolumn{1}{c}{HINT}
\\ 
\midrule[0.8pt]
\multicolumn{1}{l|}{FLOPs/G} & \multicolumn{1}{c}{6887} & \multicolumn{1}{c}{785} & \multicolumn{1}{c}{\textbf{12.00}} & \multicolumn{1}{c|}{144.25}&  \multicolumn{1}{c}{\underline{126.92}}%
\\
\multicolumn{1}{l|}{Parameters/M} & \multicolumn{1}{c}{{115.3}} & \multicolumn{1}{c}{31.76} & \multicolumn{1}{c}{\textbf{5.29}} & \multicolumn{1}{c|}{26.13}& \multicolumn{1}{c}{\underline{24.87}}
\\
\multicolumn{1}{l|}{Run-times/s}& \multicolumn{1}{c}{2.23} & \multicolumn{1}{c}{\underline{0.19}}  & \multicolumn{1}{c}{\textbf{0.13}}&\multicolumn{1}{c|}{0.29}&\multicolumn{1}{c}{0.28}
\\
\multicolumn{1}{l|}{PSNR/dB}& \multicolumn{1}{c}{18.30} & \multicolumn{1}{c}{\underline{21.94}}  & \multicolumn{1}{c}{19.66}&\multicolumn{1}{c|}{21.41}&\multicolumn{1}{c}{\textbf{27.17}}%
\\
\bottomrule[0.8pt]
\end{tabular}}
\label{tab:efficiency_compare}
\vspace{-1mm}
\end{table}

\begin{table}[t]
\caption{Quantitative comparisons (MANIQA~\cite{yang2022maniqa}) on real-world datasets. 
These results are all obtained from testing with their best LOL-v2-syn~\cite{wang2019spatial} weights. [$\uparrow$ :Higher value denotes better quality]}
\centering
\vspace{-2mm}
\scalebox{.74}
{
\begin{tabular}{ccccccc}
\toprule[0.8pt]
\multicolumn{1}{l|}{{Dataset}} & \multicolumn{1}{c}{DICM~\cite{lee2013contrast}}& \multicolumn{1}{c}{MEF~\cite{ma2015perceptual}} & \multicolumn{1}{c}{NPE~\cite{wang2013naturalness}} & \multicolumn{1}{c|}{VV~\cite{vonikakis2018evaluation}}& \multicolumn{1}{c}{Mean$\uparrow$}
\\ 
\midrule[0.8pt]
\multicolumn{1}{l|}{SNR-Net~\cite{xu2022snr}} & \multicolumn{1}{c}{{0.465}} & \multicolumn{1}{c}{0.527} & \multicolumn{1}{c}{0.480} & \multicolumn{1}{c|}{0.239}& \multicolumn{1}{c}{0.428}
\\
\multicolumn{1}{l|}{Uformer~\cite{wang2022uformer}}& \multicolumn{1}{c}{0.526} & \multicolumn{1}{c}{0.634}  & \multicolumn{1}{c}{\underline{0.515}}&\multicolumn{1}{c|}{\underline{0.356}}&\multicolumn{1}{c}{\underline{0.508}}
\\
\multicolumn{1}{l|}{Restormer~\cite{zamir2022restormer}}& \multicolumn{1}{c}{\underline{0.535}} & \multicolumn{1}{c}{\underline{0.641}}  & \multicolumn{1}{c}{0.491}&\multicolumn{1}{c|}{\underline{0.356}}&\multicolumn{1}{c}{0.507}%
\\
\multicolumn{1}{l|}{Sparse~\cite{yang2021sparse}}& \multicolumn{1}{c}{0.458} & \multicolumn{1}{c}{0.532}  & \multicolumn{1}{c}{0.324}&\multicolumn{1}{c|}{0.341}&\multicolumn{1}{c}{0.413}
\\
\midrule[0.8pt]
\multicolumn{1}{l|}{HINT} & \multicolumn{1}{c}{\textbf{0.583}} & \multicolumn{1}{c}{\textbf{0.642}} & \multicolumn{1}{c}{\textbf{0.547}} & \multicolumn{1}{c|}{\textbf{0.448}}&  \multicolumn{1}{c}{\textbf{0.555}}%
\\
\bottomrule
\end{tabular}}
\label{tab:real_world}
\vspace{-3mm}
\end{table}

\begin{figure}[htp!]
\footnotesize
\centering
\begin{tabular}{ccc}
\hspace{-0.42cm}
\begin{adjustbox}{valign=t}
\begin{tabular}{ccc}
\includegraphics[width=0.15\textwidth]{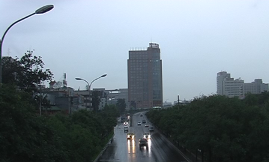} \hspace{-4mm} &
\includegraphics[width=0.15\textwidth]{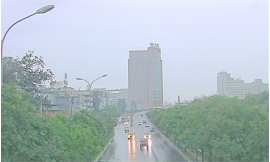} \hspace{-4mm} &
\includegraphics[width=0.15\textwidth]{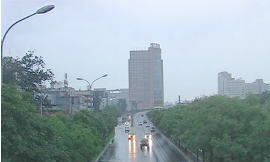} \hspace{-4mm} 
\\
 Input \hspace{-4mm} &
 Uformer~\cite{wang2022uformer} \hspace{-4mm} &
 HINT \hspace{-4mm} 
\\
\includegraphics[width=0.15\textwidth,height=2cm]{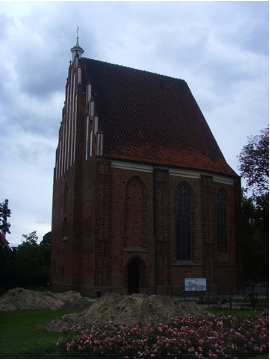} \hspace{-4mm} &
\includegraphics[width=0.15\textwidth,height=2cm]{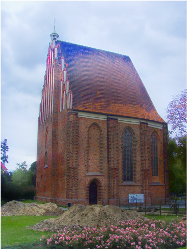} \hspace{-4mm} &
\includegraphics[width=0.15\textwidth,height=2cm]{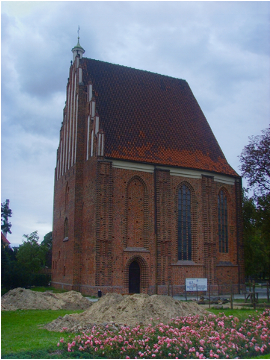} \hspace{-4mm} 
\\
 Input \hspace{-4mm} &
 Sparse~\cite{yang2021sparse} \hspace{-4mm} &
 HINT \hspace{-4mm} 
\\
\includegraphics[width=0.15\textwidth]{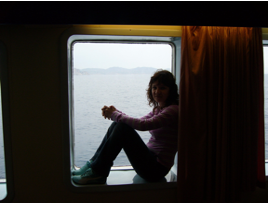} \hspace{-4mm} &
\includegraphics[width=0.15\textwidth]{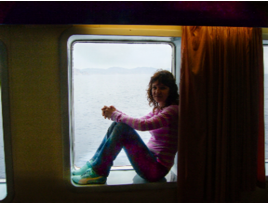} \hspace{-4mm} &
\includegraphics[width=0.15\textwidth]{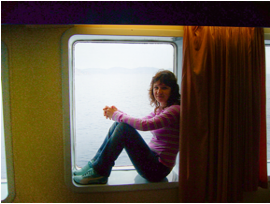} \hspace{-4mm} 
\\
 Input \hspace{-4mm} &
 SNRNet~\cite{xu2022snr} \hspace{-4mm} &
 HINT \hspace{-4mm} 
\\
\includegraphics[width=0.15\textwidth,height=3.5cm]{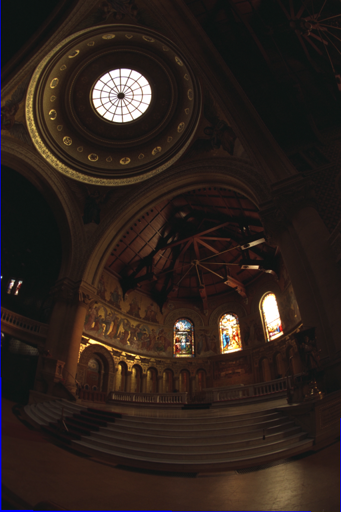} \hspace{-4mm} &
\includegraphics[width=0.15\textwidth,height=3.5cm]{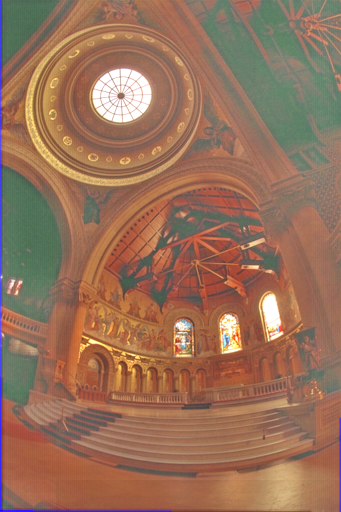} \hspace{-4mm} &
\includegraphics[width=0.15\textwidth,height=3.5cm]{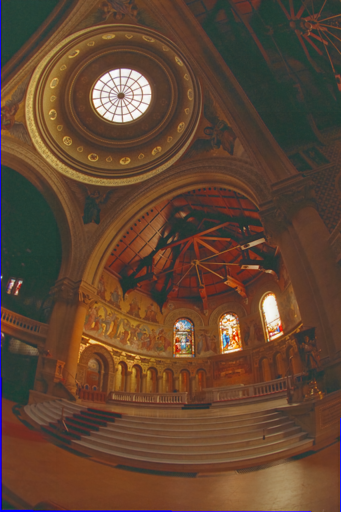} \hspace{-4mm} 
\\
 Input \hspace{-4mm} &
 Restormer~\cite{zamir2022restormer} \hspace{-4mm} &
 HINT \hspace{-4mm} 
\\
\end{tabular}
\end{adjustbox}
\end{tabular}
\vspace{-2.5mm}
\caption{Visualizations on the real-world benchmarks, including NPE~\cite{wang2013naturalness}, DICM\cite{lee2013contrast}, VV~\cite{vonikakis2018evaluation}, and MEF~\cite{ma2015perceptual} (top to bottom). 
HINT restores a pleasant result, whereas the considered technologies meet under-/over-exposed problem~\cite{zamir2022restormer,xu2022snr}, or trigger color distortion~\cite{yang2021sparse}, or remain significant artifacts~\cite{wang2022uformer}. 
}
\vspace{-4.5mm}
\label{pic:vis_realworld}
\end{figure}

\noindent\textbf{Model efficiency.}
We provide the comparison of model efficiency in terms of complexity~(FLOPs and Parameters), latency~(Run-times), and performance~(PSNR), as shown in Table~\ref{tab:efficiency_compare}. 
HINT obtains the highest PSNR score, while maintaining lower model complexity than CNN-based MIRNet~\cite{zamir2020learning}, Transformer-based IPT~\cite{chen2021pre} and Restormer~\cite{zamir2022restormer}. 

\noindent\textbf{Evaluation on real-world scenarios.} 
To assess the performance of HINT under real-world scenarios, we test it on real-world dataset without ground-truth, including DICM~\cite{lee2013contrast}, MEF~\cite{ma2015perceptual}, NPE~\cite{wang2013naturalness}, and VV~\cite{vonikakis2018evaluation}. 
The quantitative comparisons are summarized in Table~\ref{tab:real_world}, where HINT outperforms all other methods. 
%
Furthermore, as shown in Figure~\ref{pic:vis_realworld}, HINT restores a visually pleasing restoration, whereas the considered technologies meet under-/over-exposed problem~\cite{zamir2022restormer,xu2022snr}, trigger color distortion~\cite{yang2021sparse}, or remain significant artifacts~\cite{wang2022uformer}. 

\begin{figure}[tp!]
\footnotesize
\centering
\begin{tabular}{ccc}
\hspace{-0.42cm}
\begin{adjustbox}{valign=t}
\begin{tabular}{cc}
\includegraphics[width=0.2\textwidth]{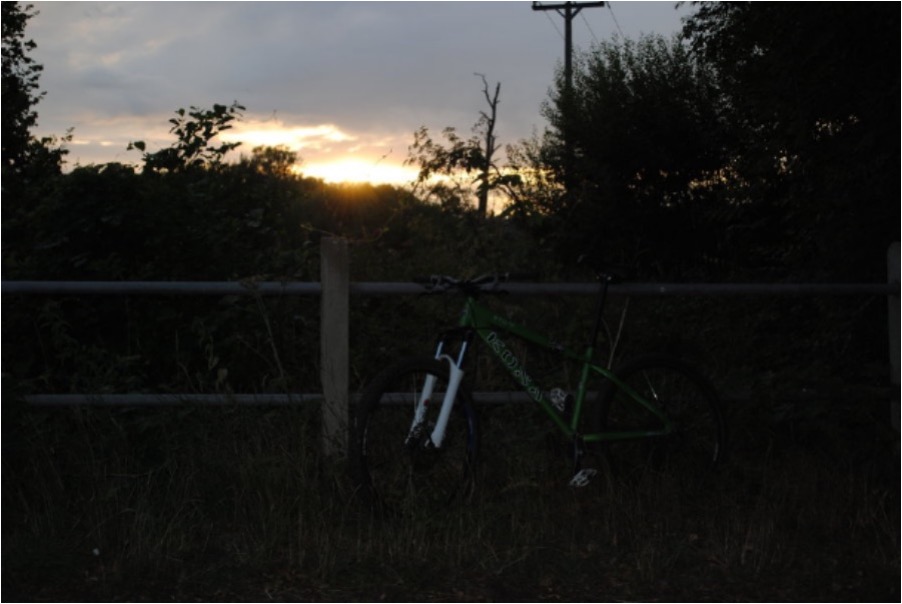} \hspace{-4mm} &
\includegraphics[width=0.2\textwidth]{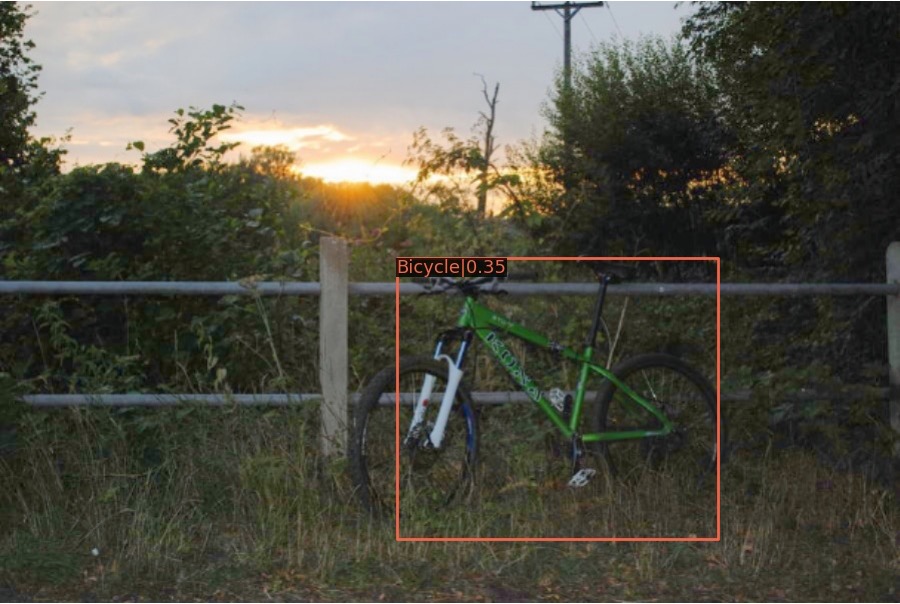} \hspace{-4mm} 
\\
 Input \hspace{-4mm} &
 HINT \hspace{-4mm} 
\end{tabular}
\end{adjustbox}
\end{tabular}
\vspace{-3mm}
\caption{Visual comparison of low-light object detection. 
Compared to the low-light input (left), the detector can predict a well-placed bounding box on the image restored by HINT (right). 
}
\vspace{-3mm}
\label{pic:vis_llod}
\end{figure}

\noindent\textbf{Application to high-level vision Task.} 
We conduct the low-light object detection experiment on the ExDark~\cite{loh2019getting} dataset. 
HINT is pre-trained on the LOL-v2-syn subset and directly applied to enhance the low-light images, with the YOLO-v3~\cite{redmon2018yolov3} model serving as the detector. 
The enhanced images bring benefits for the downstream task in qualitative (Figure~\ref{pic:vis_llod}) and quantitative ways (Table~\ref{tab:exdark_quantitatibe2OD}).

\begin{table}\footnotesize
\caption{Quantitative result of low-light object detection ExDark~\cite{loh2019getting} benchmark. 
HINT positively impacts the downstream task, and boosts the average precision (AP) scores to 7.6\%.
}
\vspace{-2.5mm}
\centering
\footnotesize
\setlength{\tabcolsep}{2.25mm}
{
\begin{tabular}{cccc}
\toprule[0.8pt]
\multicolumn{1}{l|}{Metric} & \multicolumn{1}{c}{Input}  &\multicolumn{1}{c|}{HINT} &\multicolumn{1}{c}{$\Delta$}\\ 
\midrule[0.8pt]
\multicolumn{1}{l|}{Mean (average precision, AP)} &\multicolumn{1}{c}{45.1}&\multicolumn{1}{c|}{52.7}&\multicolumn{1}{c}{+7.6}\\
\bottomrule[0.8pt]
\end{tabular}}
\label{tab:exdark_quantitatibe2OD}
\vspace{-3mm}
\end{table}

\begin{figure}[tb!]
\footnotesize
\centering
\begin{tabular}{ccc}
\hspace{-0.42cm}
\begin{adjustbox}{valign=t}
\begin{tabular}{cc}
\includegraphics[width=0.2\textwidth]{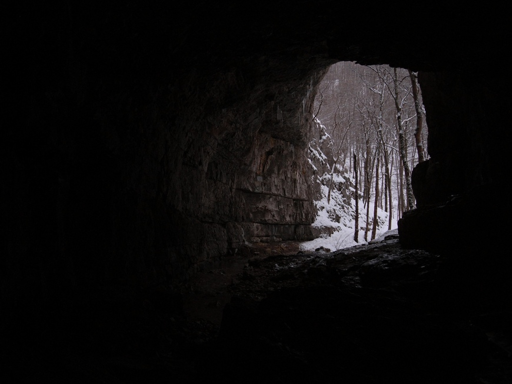} \hspace{-4mm} &
\includegraphics[width=0.2\textwidth]{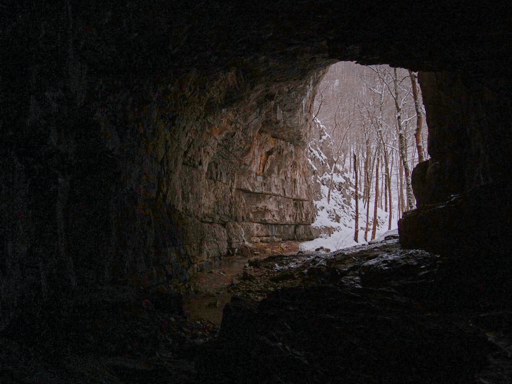} \hspace{-4mm} 
\\
 Input \hspace{-4mm} &
 HINT \hspace{-4mm} 
\end{tabular}
\end{adjustbox}
\end{tabular}
\vspace{-3mm}
\caption{Failure Case. 
HINT, which is pre-trained on the synthetic dataset, meets the challenges of restoring input under extremely low-light conditions.
}
\vspace{-5mm}
\label{pic:failure_case}
\end{figure}

%% file: sec/5_con.tex
\section{Conclusion}
\vspace{-1.mm}
\label{sec:con}
We present HINT, a Transformer-based pipeline for image restoration. 
Without bells and whistles, HINT is simple yet effective for \textbf{5} typical restoration tasks on \textbf{12} benchmarks, which demonstrates the effect of two design choices: 1) Hierarchical Multi-Head Attention (HMHA); 2)  Query-Key Cache Updating (QKCU). 
By leveraging HMHA to migrate redundancy that is rooted in the vanilla Multi-Head Attention (MHA), and introducing QKCU to enhance interactions between heads via intra- and inter- modulation, HINT is able to perform favorably against previous state-of-the-art algorithms in terms of model complexity and accuracy. 
To the best of our knowledge, HINT is the first work to restore high-quality images by exploring an efficient MHA mechanism within the widely-used Transformer architecture. 
This work provides a promising direction to achieve superior restoration performance, and we hope this community can ignite the interest and benefit from it.

\noindent\textbf{Limitations.}
While HINT offers a plausible solution for image restoration, an interesting way to achieve better results is to solve the failure case it triggers. 
As shown in Figure~\ref{pic:failure_case}, HINT struggles to recover some cases under extremely low-light conditions. 
One potential reason for this could be the domain gap between the synthetic datasets used for pre-training and real-world data.
Collecting large-scale real-world datasets for further training could be a valuable direction for future work.

%% file: HINT.bbl
\begin{thebibliography}{89}
\providecommand{\natexlab}[1]{#1}
\providecommand{\url}[1]{\texttt{#1}}
\expandafter\ifx\csname urlstyle\endcsname\relax
  \providecommand{\doi}[1]{doi: #1}\else
  \providecommand{\doi}{doi: \begingroup \urlstyle{rm}\Url}\fi

\bibitem[Ahmed et~al.(2017)Ahmed, Keskar, and Socher]{ahmed2017weighted}
Karim Ahmed, Nitish~Shirish Keskar, and Richard Socher.
\newblock Weighted transformer network for machine translation.
\newblock \emph{arXiv preprint arXiv:1711.02132}, 2017.

\bibitem[Cai et~al.(2023)Cai, Bian, Lin, Wang, Timofte, and Zhang]{retinexformer}
Yuanhao Cai, Hao Bian, Jing Lin, Haoqian Wang, Radu Timofte, and Yulun Zhang.
\newblock Retinexformer: One-stage retinex-based transformer for low-light image enhancement.
\newblock In \emph{ICCV}, 2023.

\bibitem[Chen et~al.(2021{\natexlab{a}})Chen, Wang, Guo, Xu, Deng, Liu, Ma, Xu, Xu, and Gao]{chen2021pre}
Hanting Chen, Yunhe Wang, Tianyu Guo, Chang Xu, Yiping Deng, Zhenhua Liu, Siwei Ma, Chunjing Xu, Chao Xu, and Wen Gao.
\newblock Pre-trained image processing transformer.
\newblock In \emph{CVPR}, 2021{\natexlab{a}}.

\bibitem[Chen et~al.(2022)Chen, Chu, Zhang, and Sun]{chen2022simple}
Liangyu Chen, Xiaojie Chu, Xiangyu Zhang, and Jian Sun.
\newblock Simple baselines for image restoration.
\newblock In \emph{ECCV}, 2022.

\bibitem[Chen et~al.(2024)Chen, Jiang, Pei, Sun, Wang, and Yao]{ChenJPSWY24}
Tao Chen, Xiruo Jiang, Gensheng Pei, Zeren Sun, Yucheng Wang, and Yazhou Yao.
\newblock Knowledge transfer with simulated inter-image erasing for weakly supervised semantic segmentation.
\newblock In \emph{ECCV}, 2024.

\bibitem[Chen et~al.(2020)Chen, Fang, Ding, Tsai, and Kuo]{chen2020jstasr}
Wei-Ting Chen, Hao-Yu Fang, Jian-Jiun Ding, Cheng-Che Tsai, and Sy-Yen Kuo.
\newblock Jstasr: Joint size and transparency-aware snow removal algorithm based on modified partial convolution and veiling effect removal.
\newblock In \emph{ECCV}, 2020.

\bibitem[Chen et~al.(2021{\natexlab{b}})Chen, Fang, Hsieh, Tsai, Chen, Ding, Kuo, et~al.]{Chen_2021_ICCV}
Wei-Ting Chen, Hao-Yu Fang, Cheng-Lin Hsieh, Cheng-Che Tsai, I Chen, Jian-Jiun Ding, Sy-Yen Kuo, et~al.
\newblock All snow removed: Single image desnowing algorithm using hierarchical dual-tree complex wavelet representation and contradict channel loss.
\newblock In \emph{ICCV}, 2021{\natexlab{b}}.

\bibitem[Chen et~al.(2023)Chen, Li, Li, and Pan]{DRSformer}
Xiang Chen, Hao Li, Mingqiang Li, and Jinshan Pan.
\newblock Learning a sparse transformer network for effective image deraining.
\newblock In \emph{CVPR}, 2023.

\bibitem[Cho and Lee(2009)]{cho2009fast}
Sunghyun Cho and Seungyong Lee.
\newblock Fast motion deblurring.
\newblock \emph{ACM TOG}, 28\penalty0 (5):\penalty0 1–8, 2009.

\bibitem[Cho et~al.(2021)Cho, Ji, Hong, Jung, and Ko]{cvpr2021cho}
Sung-Jin Cho, Seo-Won Ji, Jun-Pyo Hong, Seung-Won Jung, and Sung-Jea Ko.
\newblock Rethinking coarse-to-fine approach in single image deblurring.
\newblock In \emph{ICCV}, 2021.

\bibitem[Conde et~al.(2024)Conde, Geigle, and Timofte]{conde2024high}
Marcos~V Conde, Gregor Geigle, and Radu Timofte.
\newblock Instructir: High-quality image restoration following human instructions.
\newblock In \emph{ECCV}, 2024.

\bibitem[Cordonnier et~al.(2020)Cordonnier, Loukas, and Jaggi]{cordonnier2020multi}
Jean-Baptiste Cordonnier, Andreas Loukas, and Martin Jaggi.
\newblock Multi-head attention: Collaborate instead of concatenate.
\newblock \emph{arXiv preprint arXiv:2006.16362}, 2020.

\bibitem[Cui et~al.(2023{\natexlab{a}})Cui, Ren, Cao, and Knoll]{cui2023focal}
Yuning Cui, Wenqi Ren, Xiaochun Cao, and Alois Knoll.
\newblock Focal network for image restoration.
\newblock In \emph{ICCV}, 2023{\natexlab{a}}.

\bibitem[Cui et~al.(2023{\natexlab{b}})Cui, Tao, Bing, Ren, Gao, Cao, Huang, and Knoll]{cui2023selective}
Yuning Cui, Yi Tao, Zhenshan Bing, Wenqi Ren, Xinwei Gao, Xiaochun Cao, Kai Huang, and Alois Knoll.
\newblock Selective frequency network for image restoration.
\newblock In \emph{ICLR}, 2023{\natexlab{b}}.

\bibitem[Cui et~al.(2024)Cui, Ren, Cao, and Knoll]{FSNet_pami24}
Yuning Cui, Wenqi Ren, Xiaochun Cao, and Alois Knoll.
\newblock Image restoration via frequency selection.
\newblock \emph{TPAMI}, 46\penalty0 (2):\penalty0 1093--1108, 2024.

\bibitem[Cui et~al.(2025)Cui, Zamir, Khan, Knoll, Shah, and Khan]{cui2025adair}
Yuning Cui, Syed~Waqas Zamir, Salman Khan, Alois Knoll, Mubarak Shah, and Fahad~Shahbaz Khan.
\newblock Ada{IR}: Adaptive all-in-one image restoration via frequency mining and modulation.
\newblock In \emph{ICLR}, 2025.

\bibitem[Deng and Dragotti(2021)]{pami21_deng}
Xin Deng and Pier~Luigi Dragotti.
\newblock Deep convolutional neural network for multi-modal image restoration and fusion.
\newblock \emph{TPAMI}, 43\penalty0 (10):\penalty0 3333--3348, 2021.

\bibitem[Dong et~al.(2020)Dong, Liu, Zhang, Chen, and Qiao]{dong2020fd}
Yu Dong, Yihao Liu, He Zhang, Shifeng Chen, and Yu Qiao.
\newblock Fd-gan: Generative adversarial networks with fusion-discriminator for single image dehazing.
\newblock In \emph{AAAI}, 2020.

\bibitem[Dosovitskiy et~al.(2021)Dosovitskiy, Beyer, Kolesnikov, Weissenborn, Zhai, Unterthiner, Dehghani, Minderer, Heigold, Gelly, Uszkoreit, and Houlsby]{iclr2021_vit}
Alexey Dosovitskiy, Lucas Beyer, Alexander Kolesnikov, Dirk Weissenborn, Xiaohua Zhai, Thomas Unterthiner, Mostafa Dehghani, Matthias Minderer, Georg Heigold, Sylvain Gelly, Jakob Uszkoreit, and Neil Houlsby.
\newblock An image is worth 16x16 words: Transformers for image recognition at scale.
\newblock In \emph{ICLR}, 2021.

\bibitem[Duan et~al.(2024)Duan, Bai, Xie, Qi, Huang, and Tian]{DuanBXQHT24}
Kaiwen Duan, Song Bai, Lingxi Xie, Honggang Qi, Qingming Huang, and Qi Tian.
\newblock Centernet++ for object detection.
\newblock \emph{TPAMI}, 46\penalty0 (5):\penalty0 3509--3521, 2024.

\bibitem[Fergus et~al.(2006)Fergus, Singh, Hertzmann, Roweis, and Freeman]{fergus2006removing}
Rob Fergus, Barun Singh, Aaron Hertzmann, Sam~T. Roweis, and William~T. Freeman.
\newblock Removing camera shake from a single photograph.
\newblock \emph{ACM TOG}, 25\penalty0 (3):\penalty0 787–794, 2006.

\bibitem[Fu et~al.(2023)Fu, Xiao, Zhu, Liu, Wu, and Zha]{10035447}
Xueyang Fu, Jie Xiao, Yurui Zhu, Aiping Liu, Feng Wu, and Zheng-Jun Zha.
\newblock Continual image deraining with hypergraph convolutional networks.
\newblock \emph{TPAMI}, 45\penalty0 (8):\penalty0 9534--9551, 2023.

\bibitem[Gu et~al.(2019)Gu, Li, Gool, and Timofte]{gu2019self}
Shuhang Gu, Yawei Li, Luc~Van Gool, and Radu Timofte.
\newblock Self-guided network for fast image denoising.
\newblock In \emph{ICCV}, 2019.

\bibitem[Guo et~al.(2024)Guo, Li, Dai, Ouyang, Ren, and Xia]{guo2024mambair}
Hang Guo, Jinmin Li, Tao Dai, Zhihao Ouyang, Xudong Ren, and Shu-Tao Xia.
\newblock Mambair: A simple baseline for image restoration with state-space model.
\newblock In \emph{ECCV}, 2024.

\bibitem[He et~al.(2010)He, Sun, and Tang]{he2010single}
Kaiming He, Jian Sun, and Xiaoou Tang.
\newblock Single image haze removal using dark channel prior.
\newblock \emph{TPAMI}, 33\penalty0 (12):\penalty0 2341--2353, 2010.

\bibitem[Hendrycks and Gimpel(2016)]{hendrycks2016gaussian}
Dan Hendrycks and Kevin Gimpel.
\newblock Gaussian error linear units (gelus).
\newblock \emph{arXiv preprint arXiv:1606.08415}, 2016.

\bibitem[Jiang et~al.(2021)Jiang, Gong, Liu, Cheng, Fang, Shen, Yang, Zhou, and Wang]{jiang2021enlightengan}
Yifan Jiang, Xinyu Gong, Ding Liu, Yu Cheng, Chen Fang, Xiaohui Shen, Jianchao Yang, Pan Zhou, and Zhangyang Wang.
\newblock Enlightengan: Deep light enhancement without paired supervision.
\newblock \emph{TIP}, 30:\penalty0 2340--2349, 2021.

\bibitem[Kim et~al.(2024)Kim, Han, Ju, and Hwang]{0001HJH24}
Chanyoung Kim, Woojung Han, Dayun Ju, and Seong~Jae Hwang.
\newblock {EAGLE:} eigen aggregation learning for object-centric unsupervised semantic segmentation.
\newblock In \emph{CVPR}, 2024.

\bibitem[Kong et~al.(2023)Kong, Dong, Ge, Li, and Pan]{kong2023efficient}
Lingshun Kong, Jiangxin Dong, Jianjun Ge, Mingqiang Li, and Jinshan Pan.
\newblock Efficient frequency domain-based transformers for high-quality image deblurring.
\newblock In \emph{CVPR}, 2023.

\bibitem[Kupyn et~al.(2019)Kupyn, Martyniuk, Wu, and Wang]{kupyn2019deblurgan}
Orest Kupyn, Tetiana Martyniuk, Junru Wu, and Zhangyang Wang.
\newblock Deblurgan-v2: Deblurring (orders-of-magnitude) faster and better.
\newblock In \emph{ICCV}, 2019.

\bibitem[Lee et~al.(2013)Lee, Lee, and Kim]{lee2013contrast}
Chulwoo Lee, Chul Lee, and Chang-Su Kim.
\newblock Contrast enhancement based on layered difference representation of 2d histograms.
\newblock \emph{TIP}, 22\penalty0 (12):\penalty0 5372--5384, 2013.

\bibitem[Li et~al.(2017)Li, Peng, Wang, Xu, and Feng]{li2017aod}
Boyi Li, Xiulian Peng, Zhangyang Wang, Jizheng Xu, and Dan Feng.
\newblock Aod-net: All-in-one dehazing network.
\newblock In \emph{ICCV}, 2017.

\bibitem[Li et~al.(2018)Li, Ren, Fu, Tao, Feng, Zeng, and Wang]{li2018benchmarking}
Boyi Li, Wenqi Ren, Dengpan Fu, Dacheng Tao, Dan Feng, Wenjun Zeng, and Zhangyang Wang.
\newblock Benchmarking single-image dehazing and beyond.
\newblock \emph{TIP}, 28\penalty0 (1):\penalty0 492--505, 2018.

\bibitem[Li et~al.(2022)Li, Liu, Hu, Wu, Lv, and Peng]{li2022all}
Boyun Li, Xiao Liu, Peng Hu, Zhongqin Wu, Jiancheng Lv, and Xi Peng.
\newblock All-in-one image restoration for unknown corruption.
\newblock In \emph{CVPR}, 2022.

\bibitem[Li et~al.(2019)Li, Yang, Dou, Wang, Lyu, and Tu]{acl_li2019information}
Jian Li, Baosong Yang, Zi-Yi Dou, Xing Wang, Michael~R Lyu, and Zhaopeng Tu.
\newblock Information aggregation for multi-head attention with routing-by-agreement.
\newblock In \emph{NAACL}, 2019.

\bibitem[Li et~al.(2020)Li, Tan, and Cheong]{Li_2020_CVPR}
Ruoteng Li, Robby~T. Tan, and Loong-Fah Cheong.
\newblock All in one bad weather removal using architectural search.
\newblock In \emph{CVPR}, 2020.

\bibitem[Liang et~al.(2021)Liang, Cao, Sun, Zhang, Van~Gool, and Timofte]{iccv2021_swinIR}
Jingyun Liang, Jiezhang Cao, Guolei Sun, Kai Zhang, Luc Van~Gool, and Radu Timofte.
\newblock Swinir: Image restoration using swin transformer.
\newblock In \emph{ICCV Workshops}, 2021.

\bibitem[Liu et~al.(2018{\natexlab{a}})Liu, Wen, Fan, Loy, and Huang]{liu2018non}
Ding Liu, Bihan Wen, Yuchen Fan, Chen~Change Loy, and Thomas~S Huang.
\newblock Non-local recurrent network for image restoration.
\newblock In \emph{NeurIPS}, 2018{\natexlab{a}}.

\bibitem[Liu et~al.(2021{\natexlab{a}})Liu, Ma, Zhang, Fan, and Luo]{liu2021retinex}
Risheng Liu, Long Ma, Jiaao Zhang, Xin Fan, and Zhongxuan Luo.
\newblock Retinex-inspired unrolling with cooperative prior architecture search for low-light image enhancement.
\newblock In \emph{CVPR}, 2021{\natexlab{a}}.

\bibitem[Liu et~al.(2019)Liu, Suganuma, Sun, and Okatani]{liu2019dual}
Xing Liu, Masanori Suganuma, Zhun Sun, and Takayuki Okatani.
\newblock Dual residual networks leveraging the potential of paired operations for image restoration.
\newblock In \emph{CVPR}, 2019.

\bibitem[Liu et~al.(2022)Liu, Su, and Huang]{liu2022tuformer}
Xiaoyu Liu, Jiahao Su, and Furong Huang.
\newblock Tuformer: Data-driven design of transformers for improved generalization or efficiency.
\newblock In \emph{ICLR}, 2022.

\bibitem[Liu et~al.(2018{\natexlab{b}})Liu, Jaw, Huang, and Hwang]{liu2018desnownet}
Yun-Fu Liu, Da-Wei Jaw, Shih-Chia Huang, and Jenq-Neng Hwang.
\newblock Desnownet: Context-aware deep network for snow removal.
\newblock \emph{TIP}, 27\penalty0 (6):\penalty0 3064--3073, 2018{\natexlab{b}}.

\bibitem[Liu et~al.(2021{\natexlab{b}})Liu, Lin, Cao, Hu, Wei, Zhang, Lin, and Guo]{liu2021swin}
Ze Liu, Yutong Lin, Yue Cao, Han Hu, Yixuan Wei, Zheng Zhang, Stephen Lin, and Baining Guo.
\newblock Swin transformer: Hierarchical vision transformer using shifted windows.
\newblock In \emph{ICCV}, 2021{\natexlab{b}}.

\bibitem[Loh and Chan(2019)]{loh2019getting}
Yuen~Peng Loh and Chee~Seng Chan.
\newblock Getting to know low-light images with the exclusively dark dataset.
\newblock \emph{CVIU}, 178:\penalty0 30--42, 2019.

\bibitem[Luo et~al.(2021)Luo, Wu, and Guo]{nips21_luo_sr}
Fangzhou Luo, Xiaolin Wu, and Yanhui Guo.
\newblock Functional neural networks for parametric image restoration problems.
\newblock In \emph{NeurIPS}, 2021.

\bibitem[Ma et~al.(2015)Ma, Zeng, and Wang]{ma2015perceptual}
Kede Ma, Kai Zeng, and Zhou Wang.
\newblock Perceptual quality assessment for multi-exposure image fusion.
\newblock \emph{TIP}, 24\penalty0 (11):\penalty0 3345--3356, 2015.

\bibitem[Michel et~al.(2019)Michel, Levy, and Neubig]{16better1_nips19}
Paul Michel, Omer Levy, and Graham Neubig.
\newblock Are sixteen heads really better than one?
\newblock In \emph{NeurIPS}, 2019.

\bibitem[Nguyen et~al.(2022{\natexlab{a}})Nguyen, Nguyen, Do, Nguyen, Saragadam, Pham, Nguyen, Ho, and Osher]{nguyen2022improving}
Tan Nguyen, Tam Nguyen, Hai Do, Khai Nguyen, Vishwanath Saragadam, Minh Pham, Khuong~Duy Nguyen, Nhat Ho, and Stanley Osher.
\newblock Improving transformer with an admixture of attention heads.
\newblock In \emph{NeurIPS}, 2022{\natexlab{a}}.

\bibitem[Nguyen et~al.(2022{\natexlab{b}})Nguyen, Nguyen, Le, Nguyen, Tran, Baraniuk, Ho, and Osher]{MGK_icml22}
Tam~Minh Nguyen, Tan~Minh Nguyen, Dung D.~D. Le, Duy~Khuong Nguyen, Viet-Anh Tran, Richard Baraniuk, Nhat Ho, and Stanley Osher.
\newblock Improving transformers with probabilistic attention keys.
\newblock In \emph{ICML}, 2022{\natexlab{b}}.

\bibitem[Potlapalli et~al.(2023)Potlapalli, Zamir, Khan, and Shahbaz~Khan]{potlapalli2024promptir}
Vaishnav Potlapalli, Syed~Waqas Zamir, Salman~H Khan, and Fahad Shahbaz~Khan.
\newblock Promptir: Prompting for all-in-one image restoration.
\newblock In \emph{NeurIPS}, 2023.

\bibitem[Purohit et~al.(2021)Purohit, Suin, Rajagopalan, and Boddeti]{purohit2021spatially}
Kuldeep Purohit, Maitreya Suin, AN Rajagopalan, and Vishnu~Naresh Boddeti.
\newblock Spatially-adaptive image restoration using distortion-guided networks.
\newblock In \emph{ICCV}, 2021.

\bibitem[Qu et~al.(2019)Qu, Chen, Huang, and Xie]{qu2019enhanced}
Yanyun Qu, Yizi Chen, Jingying Huang, and Yuan Xie.
\newblock Enhanced pix2pix dehazing network.
\newblock In \emph{CVPR}, 2019.

\bibitem[Redmon(2018)]{redmon2018yolov3}
Joseph Redmon.
\newblock Yolov3: An incremental improvement.
\newblock \emph{arXiv preprint arXiv:1804.02767}, 2018.

\bibitem[Shazeer et~al.(2020)Shazeer, Lan, Cheng, Ding, and Hou]{shazeer2020talking}
Noam Shazeer, Zhenzhong Lan, Youlong Cheng, Nan Ding, and Le Hou.
\newblock Talking-heads attention.
\newblock \emph{arXiv preprint arXiv:2003.02436}, 2020.

\bibitem[Song et~al.(2023{\natexlab{a}})Song, Zhou, Li, Dai, Shen, Zhang, and Li]{tip23_songxibin}
Xibin Song, Dingfu Zhou, Wei Li, Yuchao Dai, Zhelun Shen, Liangjun Zhang, and Hongdong Li.
\newblock Tusr-net: Triple unfolding single image dehazing with self-regularization and dual feature to pixel attention.
\newblock \emph{TIP}, 32:\penalty0 1231--1244, 2023{\natexlab{a}}.

\bibitem[Song et~al.(2023{\natexlab{b}})Song, He, Qian, and Du]{10076399}
Yuda Song, Zhuqing He, Hui Qian, and Xin Du.
\newblock Vision transformers for single image dehazing.
\newblock \emph{TIP}, 32:\penalty0 1927--1941, 2023{\natexlab{b}}.

\bibitem[Tsai et~al.(2022)Tsai, Peng, Lin, Tsai, and Lin]{eccv2022_Stripformer}
Fu-Jen Tsai, Yan-Tsung Peng, Yen-Yu Lin, Chung-Chi Tsai, and Chia-Wen Lin.
\newblock Stripformer: Strip transformer for fast image deblurring.
\newblock In \emph{ECCV}, 2022.

\bibitem[Valanarasu et~al.(2022)Valanarasu, Yasarla, and Patel]{Valanarasu_2022_CVPR}
Jeya Maria~Jose Valanarasu, Rajeev Yasarla, and Vishal~M. Patel.
\newblock Transweather: Transformer-based restoration of images degraded by adverse weather conditions.
\newblock In \emph{CVPR}, 2022.

\bibitem[Vaswani et~al.(2017)Vaswani, Shazeer, Parmar, Uszkoreit, Jones, Gomez, Kaiser, and Polosukhin]{vaswani2017attention}
Ashish Vaswani, Noam Shazeer, Niki Parmar, Jakob Uszkoreit, Llion Jones, Aidan~N Gomez, {\L}ukasz Kaiser, and Illia Polosukhin.
\newblock Attention is all you need.
\newblock In \emph{NeurIPS}, 2017.

\bibitem[Voita et~al.(2019)Voita, Talbot, Moiseev, Sennrich, and Titov]{acl_voita2019analyzing}
Elena Voita, David Talbot, Fedor Moiseev, Rico Sennrich, and Ivan Titov.
\newblock Analyzing multi-head self-attention: Specialized heads do the heavy lifting, the rest can be pruned.
\newblock In \emph{ACL}, 2019.

\bibitem[Vonikakis et~al.(2018)Vonikakis, Kouskouridas, and Gasteratos]{vonikakis2018evaluation}
Vassilios Vonikakis, Rigas Kouskouridas, and Antonios Gasteratos.
\newblock On the evaluation of illumination compensation algorithms.
\newblock \emph{MTA}, 77:\penalty0 9211--9231, 2018.

\bibitem[\vspace{0mm} Cui et~al.(2024)\vspace{0mm} Cui, Ren, Cao, and Knoll]{convIr_pami24}
Yuning \vspace{0mm} Cui, Wenqi Ren, Xiaochun Cao, and Alois Knoll.
\newblock Revitalizing convolutional network for image restoration.
\newblock \emph{TPAMI}, 46\penalty0 (12):\penalty0 9423--9438, 2024.

\bibitem[Wang et~al.(2023)Wang, Pan, Wang, Dong, Wang, Ju, and Chen]{wang2023promptrestorer}
Cong Wang, Jinshan Pan, Wei Wang, Jiangxin Dong, Mengzhu Wang, Yakun Ju, and Junyang Chen.
\newblock Promptrestorer: A prompting image restoration method with degradation perception.
\newblock In \emph{NeurIPS}, 2023.

\bibitem[Wang et~al.(2022{\natexlab{a}})Wang, Shen, Tu, Zhuang, and Liu]{wang2022improved}
Huadong Wang, Xin Shen, Mei Tu, Yimeng Zhuang, and Zhiyuan Liu.
\newblock Improved transformer with multi-head dense collaboration.
\newblock \emph{TASLP}, 30:\penalty0 2754--2767, 2022{\natexlab{a}}.

\bibitem[Wang et~al.(2013)Wang, Zheng, Hu, and Li]{wang2013naturalness}
Shuhang Wang, Jin Zheng, Hai-Miao Hu, and Bo Li.
\newblock Naturalness preserved enhancement algorithm for non-uniform illumination images.
\newblock \emph{TIP}, 22\penalty0 (9):\penalty0 3538--3548, 2013.

\bibitem[Wang et~al.(2019)Wang, Yang, Xu, Chen, Zhang, and Lau]{wang2019spatial}
Tianyu Wang, Xin Yang, Ke Xu, Shaozhe Chen, Qiang Zhang, and Rynson~WH Lau.
\newblock Spatial attentive single-image deraining with a high quality real rain dataset.
\newblock In \emph{CVPR}, 2019.

\bibitem[Wang et~al.(2024)Wang, Yang, Fu, and Liu]{Wang_2024_CVPR}
Wenjing Wang, Huan Yang, Jianlong Fu, and Jiaying Liu.
\newblock Zero-reference low-light enhancement via physical quadruple priors.
\newblock In \emph{CVPR}, 2024.

\bibitem[Wang et~al.(2004)Wang, Bovik, Sheikh, and Simoncelli]{wang2004image}
Zhou Wang, Alan~C Bovik, Hamid~R Sheikh, and Eero~P Simoncelli.
\newblock Image quality assessment: from error visibility to structural similarity.
\newblock \emph{TIP}, 13\penalty0 (4):\penalty0 600--612, 2004.

\bibitem[Wang et~al.(2022{\natexlab{b}})Wang, Cun, Bao, Zhou, Liu, and Li]{wang2022uformer}
Zhendong Wang, Xiaodong Cun, Jianmin Bao, Wengang Zhou, Jianzhuang Liu, and Houqiang Li.
\newblock Uformer: A general u-shaped transformer for image restoration.
\newblock In \emph{CVPR}, 2022{\natexlab{b}}.

\bibitem[Wei et~al.(2018)Wei, Wang, Yang, and Liu]{Chen2018Retinex}
Chen Wei, Wenjing Wang, Wenhan Yang, and Jiaying Liu.
\newblock Deep retinex decomposition for low-light enhancement.
\newblock In \emph{BMVC}, 2018.

\bibitem[Wen et~al.(2025)Wen, Xu, Li, and Xu(ATO)]{WEN2025111033}
Yanjie Wen, Ping Xu, Zhihong Li, and Wangtu Xu(ATO).
\newblock An illumination-guided dual attention vision transformer for low-light image enhancement.
\newblock \emph{PR}, 158:\penalty0 111033, 2025.

\bibitem[Weng et~al.(2024)Weng, Yan, Tai, Qian, Yang, and Li]{weng2024mamballie}
Jiangwei Weng, Zhiqiang Yan, Ying Tai, Jianjun Qian, Jian Yang, and Jun Li.
\newblock Mamballie: Implicit retinex-aware low light enhancement with global-then-local state space.
\newblock In \emph{NeurIPS}, 2024.

\bibitem[Xiao et~al.(2024)Xiao, Meng, Li, and Yuan]{xiao2024improving}
Da Xiao, Qingye Meng, Shengping Li, and Xingyuan Yuan.
\newblock Improving transformers with dynamically composable multi-head attention.
\newblock In \emph{ICML}, 2024.

\bibitem[Xiao et~al.(2023)Xiao, Fu, Liu, Wu, and Zha]{xiao2022image}
Jie Xiao, Xueyang Fu, Aiping Liu, Feng Wu, and Zheng-Jun Zha.
\newblock Image de-raining transformer.
\newblock \emph{TPAMI}, 45\penalty0 (11):\penalty0 12978--12995, 2023.

\bibitem[Xu et~al.(2022)Xu, Wang, Fu, and Jia]{xu2022snr}
Xiaogang Xu, Ruixing Wang, Chi-Wing Fu, and Jiaya Jia.
\newblock Snr-aware low-light image enhancement.
\newblock In \emph{CVPR}, 2022.

\bibitem[Yang et~al.(2022)Yang, Wu, Shi, Lao, Gong, Cao, Wang, and Yang]{yang2022maniqa}
Sidi Yang, Tianhe Wu, Shuwei Shi, Shanshan Lao, Yuan Gong, Mingdeng Cao, Jiahao Wang, and Yujiu Yang.
\newblock Maniqa: Multi-dimension attention network for no-reference image quality assessment.
\newblock In \emph{CVPR}, 2022.

\bibitem[Yang et~al.(2021)Yang, Wang, Huang, Wang, and Liu]{yang2021sparse}
Wenhan Yang, Wenjing Wang, Haofeng Huang, Shiqi Wang, and Jiaying Liu.
\newblock Sparse gradient regularized deep retinex network for robust low-light image enhancement.
\newblock \emph{TIP}, 30:\penalty0 2072--2086, 2021.

\bibitem[Yao et~al.(2024)Yao, Xu, Guan, Huang, and Xiong]{NDR-Restore_tip24}
Mingde Yao, Ruikang Xu, Yuanshen Guan, Jie Huang, and Zhiwei Xiong.
\newblock Neural degradation representation learning for all-in-one image restoration.
\newblock \emph{TIP}, 33:\penalty0 5408--5423, 2024.

\bibitem[Yu et~al.(2022)Yu, Wang, Dong, Tang, and Loy]{yuke_pami22_car_blur_noise}
Ke Yu, Xintao Wang, Chao Dong, Xiaoou Tang, and Chen~Change Loy.
\newblock Path-restore: Learning network path selection for image restoration.
\newblock \emph{TPAMI}, 44\penalty0 (10):\penalty0 7078--7092, 2022.

\bibitem[Yue et~al.(2020)Yue, Zhao, Zhang, and Meng]{yue2020dual}
Zongsheng Yue, Qian Zhao, Lei Zhang, and Deyu Meng.
\newblock Dual adversarial network: Toward real-world noise removal and noise generation.
\newblock In \emph{ECCV}, 2020.

\bibitem[Zamir et~al.(2020)Zamir, Arora, Khan, Hayat, Khan, Yang, and Shao]{zamir2020learning}
Syed~Waqas Zamir, Aditya Arora, Salman Khan, Munawar Hayat, Fahad~Shahbaz Khan, Ming-Hsuan Yang, and Ling Shao.
\newblock Learning enriched features for real image restoration and enhancement.
\newblock In \emph{ECCV}, 2020.

\bibitem[Zamir et~al.(2022)Zamir, Arora, Khan, Hayat, Khan, and Yang]{zamir2022restormer}
Syed~Waqas Zamir, Aditya Arora, Salman Khan, Munawar Hayat, Fahad~Shahbaz Khan, and Ming-Hsuan Yang.
\newblock Restormer: Efficient transformer for high-resolution image restoration.
\newblock In \emph{CVPR}, 2022.

\bibitem[Zamir et~al.(2023)Zamir, Arora, Khan, Hayat, Khan, Yang, and Shao]{MIRNetv2}
Syed~Waqas Zamir, Aditya Arora, Salman Khan, Munawar Hayat, Fahad~Shahbaz Khan, Ming-Hsuan Yang, and Ling Shao.
\newblock Learning enriched features for fast image restoration and enhancement.
\newblock \emph{TPAMI}, 45\penalty0 (2):\penalty0 1934--1948, 2023.

\bibitem[Zhang et~al.(2018)Zhang, Zhou, Lin, and Sun]{zhang2018shufflenet}
Xiangyu Zhang, Xinyu Zhou, Mengxiao Lin, and Jian Sun.
\newblock Shufflenet: An extremely efficient convolutional neural network for mobile devices.
\newblock In \emph{CVPR}, 2018.

\bibitem[Zhang et~al.(2022)Zhang, Shen, Huang, Zhou, Rong, and Xiong]{emnlp_2022_mixture}
Xiaofeng Zhang, Yikang Shen, Zeyu Huang, Jie Zhou, Wenge Rong, and Zhang Xiong.
\newblock Mixture of attention heads: Selecting attention heads per token.
\newblock In \emph{EMNLP}, 2022.

\bibitem[Zhang et~al.(2019{\natexlab{a}})Zhang, Li, Li, Zhong, and Fu]{zhang2018residual}
Yulun Zhang, Kunpeng Li, Kai Li, Bineng Zhong, and Yun Fu.
\newblock Residual non-local attention networks for image restoration.
\newblock In \emph{ICLR}, 2019{\natexlab{a}}.

\bibitem[Zhang et~al.(2019{\natexlab{b}})Zhang, Zhang, and Guo]{zhang2019kindling}
Yonghua Zhang, Jiawan Zhang, and Xiaojie Guo.
\newblock Kindling the darkness: A practical low-light image enhancer.
\newblock In \emph{ACMMM}, 2019{\natexlab{b}}.

\bibitem[Zhou et~al.(2024{\natexlab{a}})Zhou, Chen, Pan, Shi, and Yang]{zhou2024AST}
Shihao Zhou, Duosheng Chen, Jinshan Pan, Jinglei Shi, and Jufeng Yang.
\newblock Adapt or perish: Adaptive sparse transformer with attentive feature refinement for image restoration.
\newblock In \emph{CVPR}, 2024{\natexlab{a}}.

\bibitem[Zhou et~al.(2024{\natexlab{b}})Zhou, Pan, Shi, Chen, Qu, and Yang]{zhou2025seeing}
Shihao Zhou, Jinshan Pan, Jinglei Shi, Duosheng Chen, Lishen Qu, and Jufeng Yang.
\newblock Seeing the unseen: A frequency prompt guided transformer for image restoration.
\newblock In \emph{ECCV}, 2024{\natexlab{b}}.

\end{thebibliography}
